\pdfoutput=1
\documentclass[11pt]{article}
\usepackage[margin=1in]{geometry}
\usepackage[utf8]{inputenc}
\usepackage[T1]{fontenc}
\usepackage[round,authoryear]{natbib}
\usepackage{microtype}
\usepackage{amsmath,amssymb,amsthm,mathtools,bm}
\usepackage{booktabs}
\usepackage{graphicx}
\usepackage{subcaption}
\usepackage{xcolor}
\usepackage{enumitem}
\usepackage{placeins}
\usepackage[colorlinks=true,linkcolor=blue,citecolor=blue,urlcolor=blue]{hyperref}
\usepackage[capitalize,nameinlink]{cleveref}
\graphicspath{{figs/}}
\setlist[itemize]{leftmargin=1.2em,itemsep=0.1em,topsep=0.2em}
\setlist[enumerate]{leftmargin=1.2em,itemsep=0.1em,topsep=0.2em}

\newtheorem{definition}{Definition}

\newtheorem{proposition}{Proposition}
\newcommand{\R}{\mathbb{R}}
\newcommand{\FourP}{\ensuremath{\mathsf{4P}}}
\newcommand{\ThreeNOneP}{\ensuremath{\mathsf{3N1P}}}

\newcommand{\dense}{\mathrm{dense}}

\title{Toy Combinatorial Interpretability Models Reveal Lottery Tickets in Early Feature Space}
\author{Alon Bebchuk\\Tel-Aviv University \and Nir Shavit\\MIT and Red Hat AI}
\date{}
\begin{document}
\maketitle

\begin{abstract}
The lottery ticket hypothesis posits that dense networks contain sparse subnetworks, ``winning tickets,'' that, when rewound to their initial weights and retrained in isolation, match the performance of the full model. We ask a more mechanistic question: what internal object does a winning ticket preserve? We work in a combinatorial, clause-structured toy setting that admits an interpretable feature-space representation with well-defined combinatorial distances between features. We show that winning tickets in weight space correspond to precursor locations in feature space that are already near, at initialization, to the final feature-channel codes. Dense SGD resolves these locations through structured selection: proximal locations either converge to final codes or are rejected, with rejection concentrated at more crowded neurons, implicating competition under superposition. A winning ticket is thus a family of compatible code locations that jointly balance proximity to final codes with low inter-feature interference. Sparse retraining often re-expresses the same clause/template family on a different row, so the preserved object is family-level rather than microscopic row identity. We validate this account with lightweight probes based on feature-space distance and motion; in our setting, these probes frequently outperform established weight-based ticket discovery methods in both accuracy and exact code recovery. Although these findings are grounded in a toy setting, they suggest that the lottery ticket structure is governed by hidden feature-space geometry rather than weight-space subnetwork identity.
\end{abstract}

\section{Introduction}

The lottery ticket hypothesis (LTH) is often stated in functional language: prune a trained dense network, rewind the surviving weights to their original initialization, retrain the sparse subnetwork, and the sparse model can often approach dense accuracy \citep{frankle2019lottery}. Later work sharpened that picture by showing that rewinding to slightly later dense checkpoints often helps and by studying the early stability of winning tickets under optimization, meaning that sparse subnetworks found after early training can remain trainable and connected to good dense solutions \citep{renda2020rewinding,frankle2020linear,you2020earlybird}. But those results leave open a more mechanistic question: \emph{what object is the ticket preserving?}

One answer is that the ticket preserves a lucky set of weights. Another is that it preserves a mask, or a mask together with a favorable optimization basin. We study a different possibility: in the models considered here, a ticket preserves enough of an internal feature-space scaffold for sparse training to rebuild the relevant computation. The mask is visible in weight space, because that is where pruning is applied, but the computation itself is easier to read in this feature-space view. Existing work strongly suggests that masks encode more than a set of large coordinates \citep{paul2022unmasking}. In ordinary deep networks, however, the relevant hidden representation is usually too opaque to say exactly what object the mask is preserving. We therefore use a clause-structured toy setting in which the internal feature space can be read directly.

The main story of the paper is that dense training makes certain locations in the initial feature space identifiable as useful ticket precursors. In our setting, SGD training of a DNF detector first constructs feature-channel codes: clause-local sign patterns that represent features using a small combinatorial alphabet, while multiple candidate features may share the same rows or embedded coordinates \citep{adler2025combinatorial}. A key tool at our disposal is the ability to define combinatorial distances between codes. Some candidate locations begin close in code distance to the relevant collection of locations and are amplified into exact codes during training; others that are also plausible early are rejected. The rejection is structured: it occurs disproportionately on crowded rows, indicating that dense SGD is resolving competition among candidate carriers under superposition, in the sense of multiple features sharing limited representational dimensions \citep{toy}. Thus, ticket formation is not simply local code formation and not simply magnitude growth. It is a process of selecting a compatible family of feature-channel codes.

Pruning changes this landscape. A mask is a sparse subset of coordinates in weight space, but applying that mask and rewinding does not merely remove a few displayed code locations. Because each clause-local entry is a sum over many embedded coordinates, zeroing the coordinates outside the mask changes the whole row pattern in feature space. The sparse initialization is therefore a new projected feature-space state. A winning ticket is a family of locations in this feature space that leaves enough degrees of freedom for sparse SGD to reconstruct compatible feature-channel codes.

We propose a set of constructive rules for finding tickets. These rules are diagnostic: they ask whether the properties exposed by dense SGD---distance to codes, margin, early motion, and compatibility among row carriers---are sufficient to recover part of the same support. The benchmark sections make this point sharper: feature-space rules are best at finding the emerging code structure early, while later weight magnitude becomes strong precisely because dense training has already projected much of that structure into the weights' magnitude ordering. The remaining gap is partly the translation problem: a promising row--clause code location must still be implemented by a sparse set of coordinates in weight space.

In this clause-structured setting, we make the following claims:
\begin{enumerate}[leftmargin=1.3em]
\item A lottery ticket is a set of feature-space locations that are close to the eventual feature-channel codes for the relevant clauses. The corresponding object in weight space is a mask and its initial values, not just a mask.
\item Dense SGD exposes tickets in feature space by driving selected locations toward canonical codes while rejecting other plausible locations on heavily loaded rows.
\item Pruning changes the feature-space representation. The rewound sparse state is not a visual restriction of the dense initial state; removing coordinates changes the whole row pattern.
\item Mechanism-inspired detectors based on feature-space distance and motion recover a large part of the same structure and outperform raw weight-space signals for early code prediction. Their success supports distance and motion as crucial ticket elements, while their limitations reveal a separate site-to-weights translation problem.
\end{enumerate}

\paragraph{Limitations}
 We are reminded of a story about Judah Folkman, the father of anti-angiogenesis therapy for cancer, a technique that today serves millions and generates billions in revenue. Some 30 years ago, after discovering the first two angiogenesis-inhibiting drugs that cured cancer in laboratory mice, a reporter asked what this astonishing result meant for cancer research. Folkman said: “Well, it says that if you are a mouse and you have cancer, we can probably cure you.” Our claim here is in that spirit: if you are a small neural network computing Boolean formulae, then we can probably understand your lottery tickets. Whether, when, and how these ideas extend to larger models and less discrete tasks is a research program. The paper should be read as a model-organism study that makes feature-space measurement possible and suggests a new path in interpretability research, not as a completed theory of lottery tickets in general.


\section{Related Work}
\label{sec:related-work}

The original lottery ticket hypothesis showed that dense networks can contain sparse subnetworks whose \emph{original initialization} allows them to train to dense-like performance when isolated and retrained \citep{frankle2019lottery}. Subsequent work established rewinding and fine-tuning as practical routes to ticket recovery and emphasized that tickets become more trainable after a small amount of dense training \citep{renda2020rewinding,frankle2020linear}. This later view also weakens a purely initialization-centric interpretation of tickets: early training can move a network into a more stable region of optimization, after which the sparse subnetwork is better understood as preserving access to a compatible trajectory or basin rather than only preserving lucky coordinate values \citep{frankle2019stabilizing,frankle2020linear}. Our paper studies a different question. Rather than asking only whether a ticket works, we ask what internal representational object it preserves.

A second line of work showed that ticket masks can become visible early in training, both through early-bird ticket heuristics and through initialization-time or near-initialization mask search \citep{you2020earlybird,savarese2020continuous,sreenivasan2022raregems,paul2022datadiet}. These papers are directly relevant to our probe experiments. We therefore compare mechanism-inspired ticket selection rules based on distances in feature space to checkpoint magnitude, SNIP, GraSP, SynFlow, and an Early-Bird-style magnitude baseline in Sections~\ref{sec:probe-rules} and~\ref{sec:broad-feature-weight-sweep}. These baselines typically score weights, gradients, or mask stability. By contrast, our ticket selection ranks \emph{feature-space locations} first and turns those locations into a ticket in weight space only afterward.

The closest conceptual prior is \citep{paul2022unmasking}, who argue that winning-ticket masks encode more than a set of large coordinates and interpret masks geometrically in terms of a favorable low-error subspace. Work in reinforcement learning also suggests that IMP can reveal sparse, task-relevant representations \citep{vischer2022minimal}. Relatedly, the generalized lottery ticket hypothesis argues that the relevant notion of sparsity need not be tied to the canonical parameter basis, and that tickets can be understood after changing the basis in which parameters are represented \citep{alabdulmohsin2021generalized}. Our work is complementary: rather than only changing the parameter basis, we define an interpretable feature-space object and ask how its support is realized by a sparse weight-space mask. Our results are complementary but more explicit: because the hidden representation is directly visualizable, we can identify the recovered object as a family of code locations, define combinatorial distances to that family, and derive a forward-looking support rule from that geometry.

Other work shows that masks and signs themselves can already carry substantial information. Supermask results and randomly weighted subnetworks weaken the naive view that useful computation must always be stored in trained weights \citep{zhou2019deconstructing,ramanujan2020hidden}. Our contribution is not to claim that support alone always suffices. It is to identify, in a setting where the internal representation is visible, what a useful support corresponds to: the weight-space realization of a ticket via corresponding feature-space code locations.

Mechanistic interpretability work on superposition studies how neural networks represent more features than available dimensions by packing features into shared representational space \citep{toy,adler2024,chan2024superposition}. This perspective is closely related to the row-level competition we observe: several clause-code precursors may compete for the same hidden row, and successful tickets select mutually compatible locations rather than simply the individually strongest coordinates. Our setting differs in that the relevant feature families are known in advance, allowing us to measure distances to canonical code families and to track which candidate locations are recruited or rejected during sparse retraining.

Mechanistic analyses of grokking show that delayed generalization corresponds
to the gradual amplification of a sparse interpretable circuit followed by a
rapid cleanup phase in which competing representations are
suppressed~\citep{nanda2023progress}.
Recent work connects this directly to lottery tickets: sparse subnetworks
identified during the generalizing phase drastically reduce delayed
generalization, suggesting that ticket selection and generalization-circuit
formation are two views of the same process~\citep{davies2024lottery}.
Our feature-space account adds precision to what that selection preserves:
a compatible family of code locations whose combinatorial geometry is already
present at initialization.

Sparse autoencoders (SAEs) provide another route to feature-space interpretability. In that line of work,
one trains an auxiliary sparse dictionary to decompose dense activations into more interpretable
features, often treating the learned decoder directions as candidate feature directions
\citep{Cunningham2024,scaling_monosemanticity_2024,anthropic2023autoencoder}. Our setting is
complementary. We do not learn a post-hoc dictionary over an opaque activation space. Instead, the toy task is constructed so that the relevant feature-space representation can be read directly and the local code families have an explicit combinatorial geometry. Thus
SAEs and our model share the goal of making superposed features visible, but our contribution is to
study lottery tickets in a setting where feature locations, code distances, and sparse-mask effects can
be measured without first solving a dictionary-learning problem.

The toy combinatorial interpretability framework we use here was introduced by \citep{adler2025combinatorial}. The use of this framework for analyzing sparse computation is due to \citep{kong2026the}. Our work adds to the framework the idea of using combinatorial distances among codes to measure distances in feature space and track the emergence of feature-channel codes during training with SGD. 

\section{A feature-space view of the toy DNF model}

\label{sec:methodology}

\begin{figure}[t]
\centering
\includegraphics[width=0.99\linewidth]{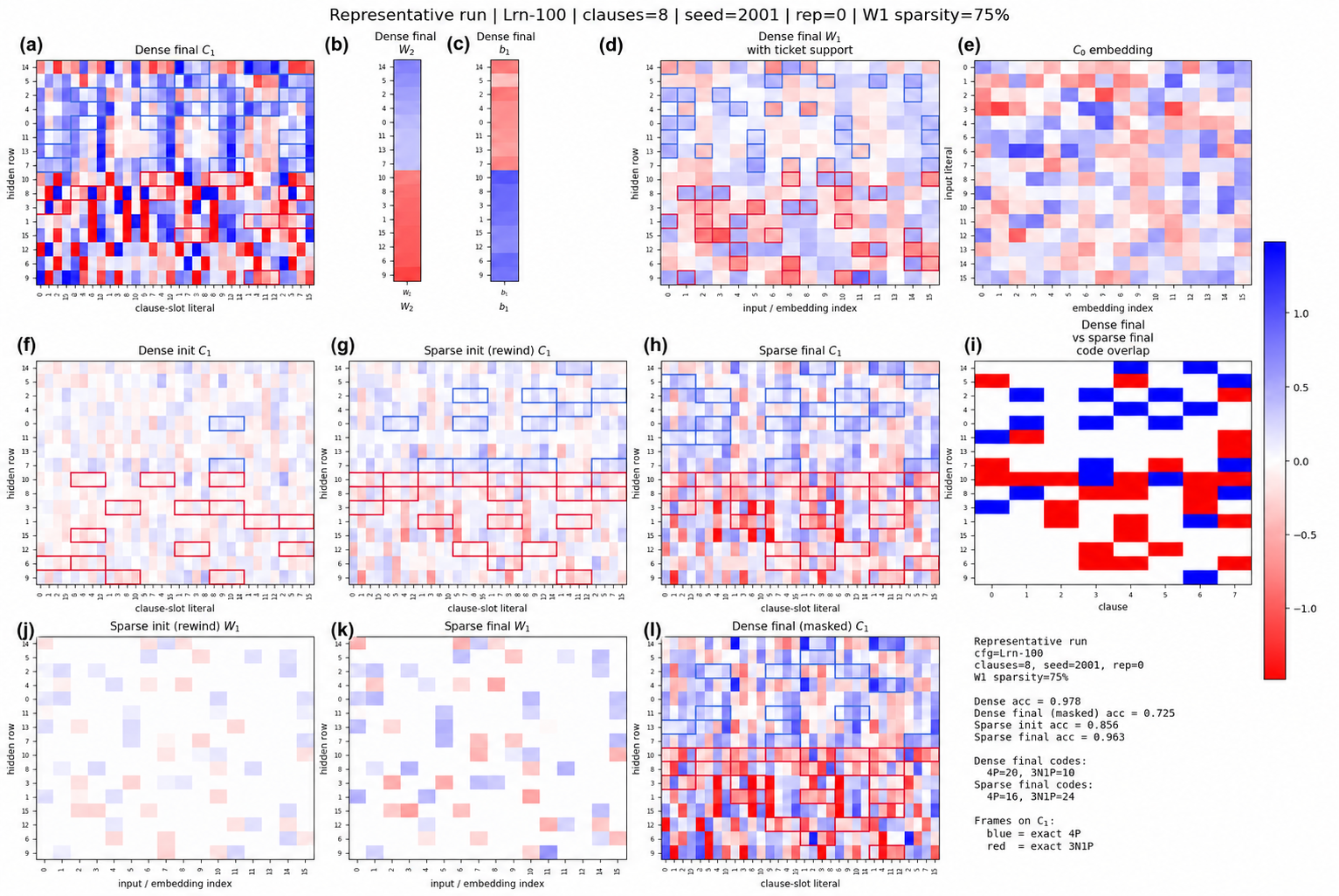}
\caption{
\textbf{Initial explanation of the setup and ticket cycle.}
The figure follows one representative run through dense training, masking, rewind, and sparse retraining.
}
\label{fig:setup}
\end{figure}

We model a single layer of a network using a canonical setup, with an added embedding layer as used in \citep{adler2025combinatorial,kong2026the}. Formally, given inputs $\mathbf{x} \in \mathbb{R}^{d_{in}}$, we first apply a learnable linear projection $\mathbf{C}_0 \in \mathbb{R}^{d \times d_{in}}$ to obtain a transformed representation $\mathbf{x'} = \mathbf{C}_0\mathbf{x}$, simulating prior layers in a deep network (If one were to put a sparse autoencoder \cite{anthropic2023autoencoder} between this layer and  the previous one, then $C_0$ would represent its decoder matrix). The network then computes output logits $\mathbf{z}$ via $\mathbf{z} = \mathbf{W}_2 \cdot \text{ReLU}(\mathbf{W}_1 \mathbf{x'} + \mathbf{b}_1) + \mathbf{b}_2$, where $\mathbf{W}_1 \in \mathbb{R}^{h \times d}$ and $\mathbf{W}_2 \in \mathbb{R}^{C \times h}$ are the weight matrices for the hidden and output layers, respectively, and $\mathbf{b}_1, \mathbf{b}_2$ are biases. To isolate the effect of feature-weight alignment, we avoid introducing dimensionality changes: we use a square linear embedding $\mathbf{C}_0$ whose output dimensionality matches its input dimensionality and set the width of $\mathbf{W}_1$ to the same value (details in Appendix~\ref{sec:training}). An example final learned state of this model is shown in the top row of Figure~\ref{fig:setup}. 

\begin{figure}[h]
  \centering
  \includegraphics[width=0.96\textwidth]{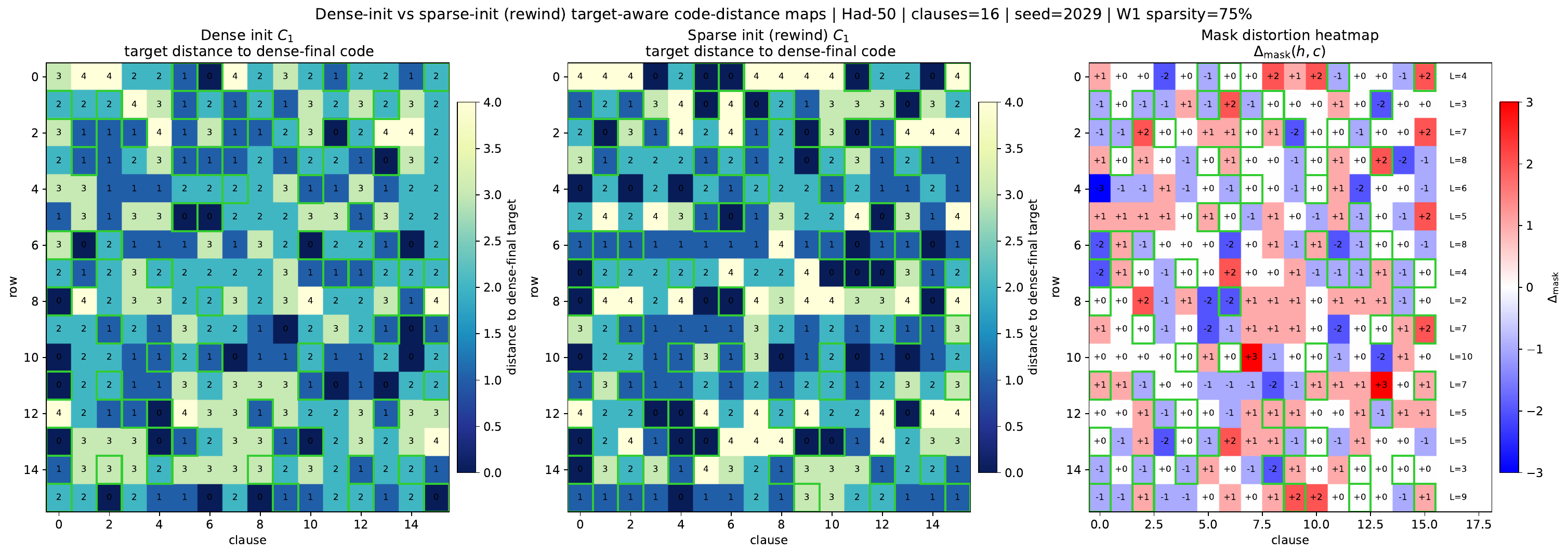}
  \caption{\textbf{The ticket support changes the feature-space initial condition.}}
  \label{fig:dense-init-sparse-init-delta-mask}
\end{figure}

$\mathbf{C}_0$ is central to our design. \citep{adler2025combinatorial,kong2026the} showed that the network is implicitly learning $\mathbf{W}_1 = \mathbf{C}_1\mathbf{C}_0^{-1}$, where $\mathbf{C}_1$ represents the feature space computation: $\mathbf{W}_1$ ``disentangles'' the relevant features from $\mathbf{x'}$ by inverting $\mathbf{C}_0$, then performs the computation in feature-space. We test cases where $C_0$ is a fixed Hadamard matrix, a random matrix, or a learned one. This allows us to test how pruning $\mathbf{W}_1$ affects the underlying feature space: $\mathbf{C}_1 = \mathbf{W}_1 \mathbf{C}_0 \in \mathbb{R}^{h \times d_{in}}$. Thus, $\mathbf{C}_1$ will be the primary object of analysis for interpreting internal representations under pruning, distinguishing functional structure in $\mathbf{C}_1$ from sparsity in $\mathbf{W}_1$ (\cref{fig:dense-init-sparse-init-delta-mask}). 

We study read-once monotone DNF tasks with clause size four. A positive-output row contributes evidence for a clause when all four literals are active. Thus an all-positive local pattern, \FourP{}, is the natural positive row code, a soft logical AND of variable $X_0 \wedge \ldots \wedge X_3$ emerging via SGD as $\mbox{ReLU}(X_0+\ldots+X_3-b)$ (if $b=1$ it would be an exact equivalent). Rows with negative output weight expose the complementary family, \ThreeNOneP{}, one positive and three negative, which supplies a negative-output clause-local code. These are not arbitrary labels, as shown in \citep{adler2025combinatorial,kong2026the}: they are the local codes through which this model expresses positive and negative evidence for the DNF. They are, however, the \emph{canonical task-aligned} code families rather than an exhaustive list of exact local sign patterns in $C_1$: trained sparse models can also contain noncanonical patterns such as $3P1N$, $2P2N$, and $4N$. Later, when we count such mixed patterns, we treat them as additional structure in the feature-space scaffold, not as replacements for the row-compatible \FourP{} and \ThreeNOneP{} evidence channels.

Let $T_{4P}=\{(+1,+1,+1,+1)\}$ and $T_{3N1P}=\{(+1,-1,-1,-1),\ldots,(-1,-1,-1,+1)\}$.
For a local vector $u\in\R^4$, define
\begin{equation}
 d_\tau(u,T)=\min_{t\in T}\sum_{r=1}^{4}\mathbf 1\{t_ru_r<\tau\},\qquad
 m(u,T)=\max_{t\in T}\min_r t_ru_r.
\end{equation}
Distance zero means an exact margin code; distance one means a one-defect near-code. The margin $m$ distinguishes a weak correct sign pattern from a stable code. 

In the rest of this paper, for the sake of clarity, most of the examples will be for a network with $W_1$ having 16 or 32 hidden neurons, 8 or 16 clauses, and a Hadamard or learned embedding. We have extensively tested various combinations of other settings and scalability of the results beyond 16 or 32 neurons and report on some of them in Appendix~\ref{sec:scaling-lottery-ticket}. 

We use a deliberately light terminology. A \emph{location} is a row--clause site $(h,c)$ in the feature-space clause-local representation $C_1$. A \emph{code} is a local clause pattern at such a location, such as a \FourP{} or \ThreeNOneP{} pattern. A \emph{feature-channel code} for a clause/template family is the collection of locations carrying that code across rows. A \emph{family} is the same clause/template identity after forgetting the carrier row. Following \citep{adler2025combinatorial}, the feature-channel code is therefore not just a single \FourP{} or \ThreeNOneP{} patch; it is the distributed placement of such codes across neurons. We use \emph{scaffold} for the pre-training or early-training collection of row--clause sites that
are already close, in the distance $d_\tau$, to a relevant code family but have not necessarily reached
positive-margin exactness. Thus a scaffold is not a finished circuit; it is a set of feature-space
precursors that sparse or dense SGD may sharpen, move, or reject.

Figure~\ref{fig:setup} is a visual guide to the rest of the paper, so lets spend some time on it. The figure shows a representative run in both weight space and feature space, where blue is positive, red negative, and zero white. The figure follows the ticket cycle from dense final, to masked final,
to dense initialization, to sparse rewind, and finally to sparse retraining.

In the top row, panels~(a)--(e), we see the final state of a model trained with SGD, where all rows are ordered by the $W_2$ contribution in panel~(b), which is itself reordered so positive rows (neurons contributing a positive signal to the DNF detection, are on top. We see no structure in $W_1$ or $C_0$ in panels~(d) and~(e), but when looking at $C_1$ in panel~(a), which is ordered by clause on the $X$ axis, 4 entries per DNF clause, we see patterns of \FourP{} have emerged in the positive rows and patterns of \ThreeNOneP{} in the negative rows (framed with blue or red frames). The $X$ axis of $C_0$ in panel~(e) is of width 32 to accommodate the 4 DNF clauses, some of which share literals with others. In $W_1$ in panel~(d) we also mark the 25\% highest magnitude weights in the final dense trained matrix that form the \textit{mask} part of the ticket \citep{frankle2019lottery}. A first crucial point to note is that while the codes in the {\it dense final} $C_1$ in panel~(a) are the result of multiplying and adding the complete rows of $W_1$ by those of $C_0$, if we look at the feature space representation of the {\it dense final (masked)} $C_1$ in panel~(l), the one resulting from using only masked locations with their dense values, it has a different feature space representation in which it has less codes, and its accuracy has dropped from $0.978$ to $0.725$. 

If we now take a look at the {\it dense init} $C_1$ state in panel~(f) we see what feature space looked like before dense training started, when the state was the result of a random dense $W_1$. This is the state we would ideally like to choose the lottery ticket from. One can see that it has many fewer \FourP or \ThreeNOneP codes and in fact as we will see many other locations in the final ticket mask are far from their final codes. 

Now comes the shocker. Look at what happens when we apply the mask to this state leaving only 25\% of the weights (the {\it sparse init (rewind)} $W_1$ in panel~(j)). We see that the number of codes suddenly goes up in the {\it sparse init (rewind)} $C_1$ in panel~(g), and accuracy jumps to $0.856$. Applying the final mask to the initial state creates a lot of codes simply by removing weights from $W_1$. This is the feature space explanation of the {\it mask-1} discovery in the supermask paper by \citep{zhou2019deconstructing}. As we can see, removal of the 75\% smaller weights has created a {\it sparse init (rewind)} $C_1$ state in panel~(g) with many more codes! 

Finally, we can look at {\it sparse final} $W_1$ and $C_1$ in panels~(k) and~(h) and we see that after training sparsely, by simply adjusting the 25\% mask weights, the network adds many more codes, and accordingly accuracy goes up from $0.856$ to $0.963$. As we will see, the mask has picked locations in the initial state that when trained will end in good feature channel codes. But this is not the whole story. If we look for example at neuron 2 (the third row from the top in $C_1$), we see that had the rightmost code for clause $\{2,5,7,12\}$ stayed \FourP\  as it was in {\it sparse init (rewind)} $C_1$ in panel~(g), that neuron would be overloaded with four \FourP\ codes in {\it sparse final} $C_1$ in panel~(h). Instead, the code is moved by training to neuron $5$ (the second row from the top) so as to leave neuron $2$ less loaded with three codes. Of course, the choices here are not straigtforward, and SGD is solving a global superpositioning of codes to allow all clauses to have sufficient signal come through without interference. The thing to note though, as we will show, is that the initial state is just a basin from which to start the training, and one could end up in many different codes cominations. In particular, the overlap figure on the middle right in panel~(i) (blue is a match, red mismatch) shows that two collections of codes, one in dense final (not masked, as in top right corner heat map in panel~(a)) feature space and the other in sparse final feature space, can have similar high accuracy by arriving at completely different codes. This is the lottery ticket hypothesis at work.

\section{Feature Space Lottery Tickets}
\label{sec:feature-space-tickets}
We now state the ticket notion used in this model. The object we call a lottery ticket is defined in the initial feature space: it is a set of row--clause locations. A weight-space mask is a witness that this feature-space object can be realized after rewind. Let $\theta_0=(W_1^0,W_2^0,\ldots)$ be the dense initialization. A mask $M$ on $W_1$ induces the rewound sparse feature representation
$C_1^M(\theta_0)=(M\odot W_1^0)C_0$,
which is not a visual restriction of $C_1^0=W_1^0C_0$. Each surviving coordinate contributes an embedding-column pattern to the whole row. Zeroing the complement of the support can improve or worsen distance to the reference code family at every location on that row. In Figure~\ref{fig:dense-init-sparse-init-delta-mask}, the left panel shows the distance from dense initialization to each location's dense-final target family. The middle panel shows the corresponding distances after applying the oracle support and rewinding. The right panel shows the distortion caused by masking. The target family map is the same in all panels. Thus pruning is not only a weight-space operation: it changes the whole initial feature-space row, and a ticket should not be described as a finished sparse circuit already sitting in the initial weights.

Fix a dense trained reference model and let $\mathcal G^\star$ denote its feature-channel family map: $(c,t)\in\mathcal G^\star$ when some dense-final row carries template $t$ for clause $c$. This target is row-forgetting: it records dense-final clause/template families, not dense-final row identities. In the experiments below, $\mathcal G^\star$ is always this dense-final family map. We also fix a visibility resolution $(r,\eta)$, where $r$ is the allowed sign-defect radius and $\eta$ is the minimum contribution score. We suppress this dependence in prose when it is clear.

For a mask $M$, write $q_M(h,c,t)$ for the contribution score of the surviving weights in row $h$ toward template $t$ on clause $c$. A location $(h,c)$ is visible under $M$ at rewind if there exists a template $t$ with $(c,t)\in\mathcal G^\star$ such that
    $d_\tau(C_1^M(\theta_0)[h,c],t)\le r
    \qquad\text{and}\qquad
    q_M(h,c,t)\ge \eta $.
Let $\Phi^{\mathcal G^\star}_{r,\eta}(M;\theta_0)$ be the set of visible locations. This visibility map is the bridge from weight space back to feature space.

\begin{definition}[Feature-space lottery ticket]
Fix $\theta_0$, a target family map $\mathcal{G}^\star$, and a visibility
resolution $(r,\eta)$. Let $E_{\dense}$ be the test error of the dense model
after training. A set of feature-space locations $P\subseteq[H]\times[K]$ is
an $(S,\epsilon)$ \emph{feature-space lottery ticket} at resolution $(r,\eta)$
if there exists a first-layer mask $M\in\{0,1\}^{H\times D}$ satisfying
\[
  \|M\|_0 \le S\|W_1^0\|_0, \qquad
  \Phi^{\mathcal{G}^\star}_{r,\eta}(M;\theta_0) = P, \qquad
  E\!\left(\mathrm{Train}(\theta_0;M)\right) \le E_{\dense}+\epsilon.
\]
The set of all such masks is the \emph{witness set}
\[
  \mathcal{W}^{\mathcal{G}^\star,r,\eta}_{S,\epsilon}(P;\theta_0),
\]
which we usually write as $\mathcal{W}_{S,\epsilon}(P;\theta_0)$ when
$\mathcal{G}^\star$, $r$, and $\eta$ are fixed. Here
$\mathrm{Train}(\theta_0;M)$ means sparse retraining from the rewound
initialization with first-layer weights $M\odot W_1^0$, while keeping the
first-layer mask fixed. Any $M$ in the witness set is a
\emph{weight-space witness} for the ticket $P$.
\end{definition}

Thus the ticket is the feature-space object $P$; a mask is only one possible sparse weight-space realization of that object. The same ticket may have several weight-space witnesses, because different sparse subsets of $W_1$ can induce the same visible locations in $C_1^M(\theta_0)$. Conversely, a mask is useful here only insofar as it witnesses a feature-space ticket whose visible locations can be refined by sparse SGD into the target feature-channel families. This is an existential definition, but the empirical sections use constructive witnesses: the relevant $P$ is induced by explicit masks, most often the dense-final oracle mask or a diagnostic feature-space rule. Degenerate location sets are not the object of study; what matters empirically is whether the induced ticket has code-family content and trains to near-dense error.

The use of $\mathcal G^\star$ should not be read as saying that the dense-final code map is the unique route to a good sparse computation. It is a fixed dense-reference family map that makes the visibility operator measurable. The lottery-ticket condition itself is functional, through the comparison to dense-model error. For this reason, later diagnostics also ask self-conditioned questions, such as whether the masked initial state is close to the sparse model's own final codes, because sparse retraining can realize a high-accuracy computation with a different compatible code family.

We use \emph{scaffold} informally for the near-code collection before it has become exact. Because $\mathcal G^\star$ is row-forgetting, the definition targets dense-final clause/template families rather than dense-final row identities. The next subsection explains why recovery should also be compared at the family level: sparse training can preserve the feature-channel code while moving a local code to a different row.

\subsection{The recovered object is a family, not a fixed row}
\label{sec:family-not-site}

Sparse retraining reveals that exact row identity is too strict as a notion of recovery. A dense-final code at location $(h,c)$ has a carrier row $h$, a clause $c$, and a template: \FourP{} for a positive row or one of the \ThreeNOneP{} templates for a negative row. Same-site recovery asks whether the sparse final model recovers the code on the same row. Family recovery forgets the row and asks whether the same clause/template family is recovered somewhere. If $P$ is a set of coded locations with fitted templates $t_{h,c}$, we write
$\mathrm{fam}(P)=\{(c,t_{h,c}) : (h,c)\in P\}$. For \FourP{} the template is unique; for \ThreeNOneP{} it records which literal is positive.

\begin{figure}[t]
  \centering
  \includegraphics[width=0.99\textwidth]{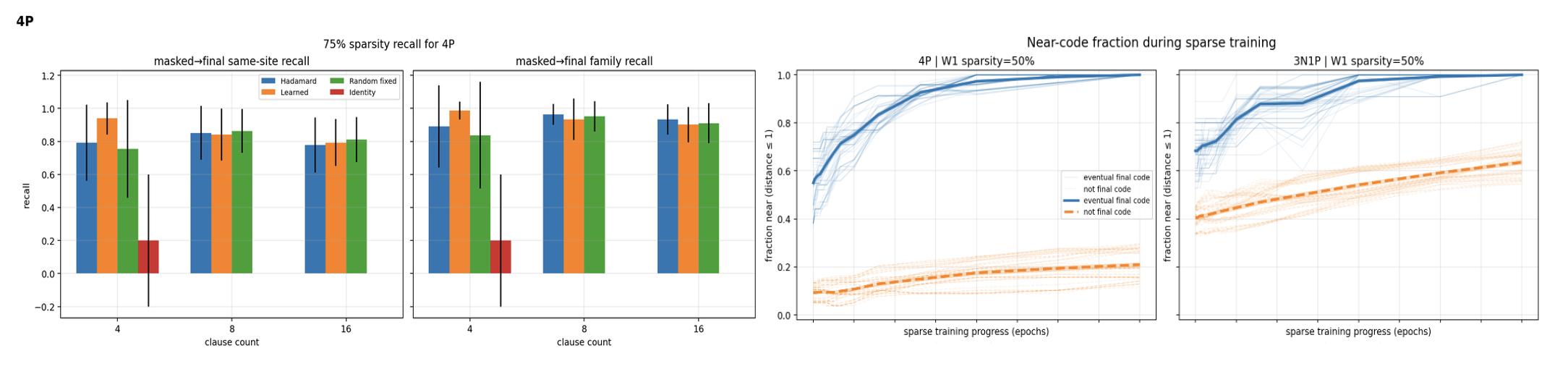}
  \caption{\textbf{Feature-space tickets are family-level and already present at rewind.} Left pair: same-site recall asks whether a dense-final \FourP{} code is recovered on the same row, while family recall forgets the row and asks whether the same clause/template family is recovered somewhere in the sparse final model. Right pair: during sparse training, locations that eventually become final codes have much higher near-code fraction from the rewound initial state and contract rapidly toward exact codes for both \FourP{} and \ThreeNOneP{}.}
  \label{fig:family-vs-site-recall}
\end{figure}

The left pair of Figure~\ref{fig:family-vs-site-recall} shows why same-site recovery is the wrong microscopic criterion. The first panel asks for literal row reuse; the second panel forgets the row and asks only whether the same clause/template family is recovered somewhere. The family-recall bars are consistently higher than the same-site bars across the distributed embeddings and clause counts. This is the direct evidence for the definition above: the useful object is not a fixed list of row identities, but a feature-space ticket whose code families can be re-expressed on compatible rows.

\subsection{Sparse retraining contracts privileged rewound precursors}
\label{sec:sparse-retraining-dynamics}

For each sparse run we group locations by whether they eventually become sparse-final exact codes. We then track, during sparse training, the fraction of locations within one sign defect of the relevant final code. The right pair of Figure~\ref{fig:family-vs-site-recall} shows the result. The blue curves are locations that eventually become final codes; already at the rewound sparse initialization they have much larger near-code fraction. The orange curves are locations that do not become final codes; they start lower and improve much more slowly. Sparse training then drives the blue curves rapidly toward near-code fraction one, separately for \FourP{} and \ThreeNOneP{}.

Thus the sparse ticket is not inventing the representation from scratch. The mask-induced rewound state already places a subset of locations in a privileged feature-space basin, and sparse SGD contracts those locations into exact codes. Together, the two halves of Figure~\ref{fig:family-vs-site-recall} give the proof-of-mechanism in this toy setting: the left pair shows that recovery is family-level rather than row-identity-level, and the right pair shows that the recovered locations were already present as near-code precursors at rewind.

\begin{figure}[thb]
\centering
\includegraphics[width=0.96\linewidth]{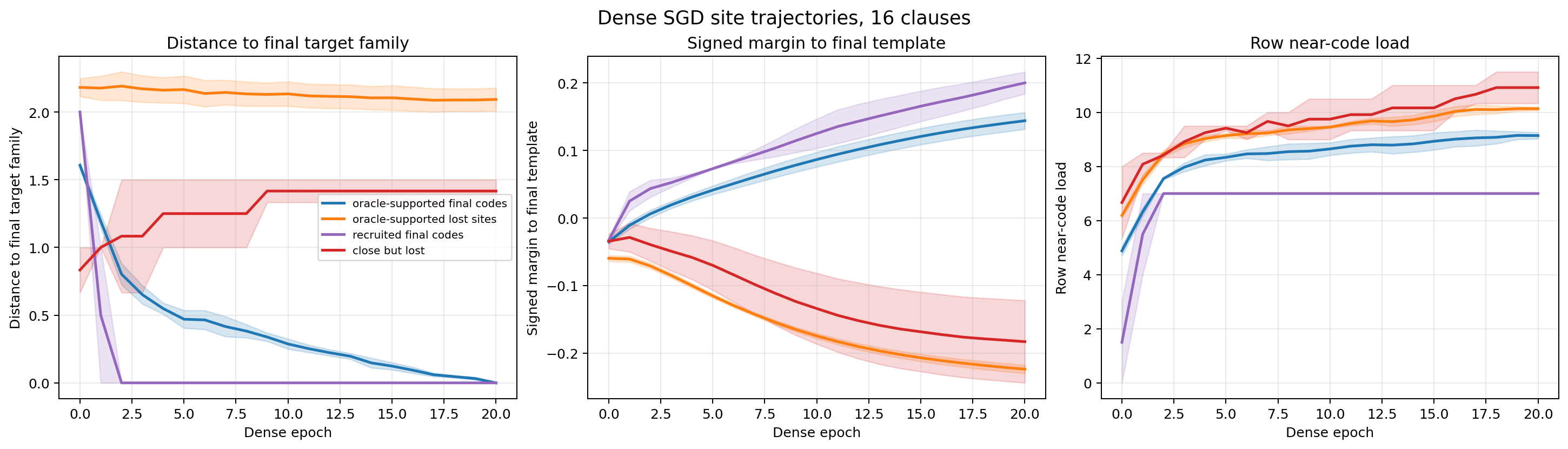}
\caption{\textbf{Supported locations split into stable codes and lost locations.} Dense training moves some precursor locations into exact codes, recruits others, and rejects locally plausible alternatives.}
\label{fig:site-trajectories}
\end{figure}

\section{Dense SGD makes code precursors legible}
\label{sec:dense-sgd-selection}

We call the magnitude-pruned mask found after dense training the \emph{oracle} mask. Dense SGD makes the eventual ticket visible in two ways. Locally, it moves selected precursor locations toward canonical codes. Globally, it rejects other initially plausible locations, especially on crowded rows, and sometimes re-expresses the same clause/template family on a different row.

Figure~\ref{fig:site-trajectories} separates locations into four categories. Oracle-supported final codes move steadily to distance zero and positive margin. Oracle-supported lost locations are also supported by the oracle mask but do not become exact codes. Recruited final codes are not obvious oracle-supported candidates early, yet rapidly become exact codes. Close-but-lost locations are the negative control: they begin locally promising but are rejected. Thus local proximity is necessary but not sufficient. SGD is solving a placement problem in feature space, not merely pushing every near-code location toward an exact code. Recruited codes are not counterexamples to the scaffold view: they show that the scaffold is
family-level and dynamical rather than a fixed list of initially exact row identities. Some families are
made available by the support but become exact only after other rows reject or release competing
candidate locations.

\begin{figure}[thb]
    \centering
    \includegraphics[width=0.98\linewidth]{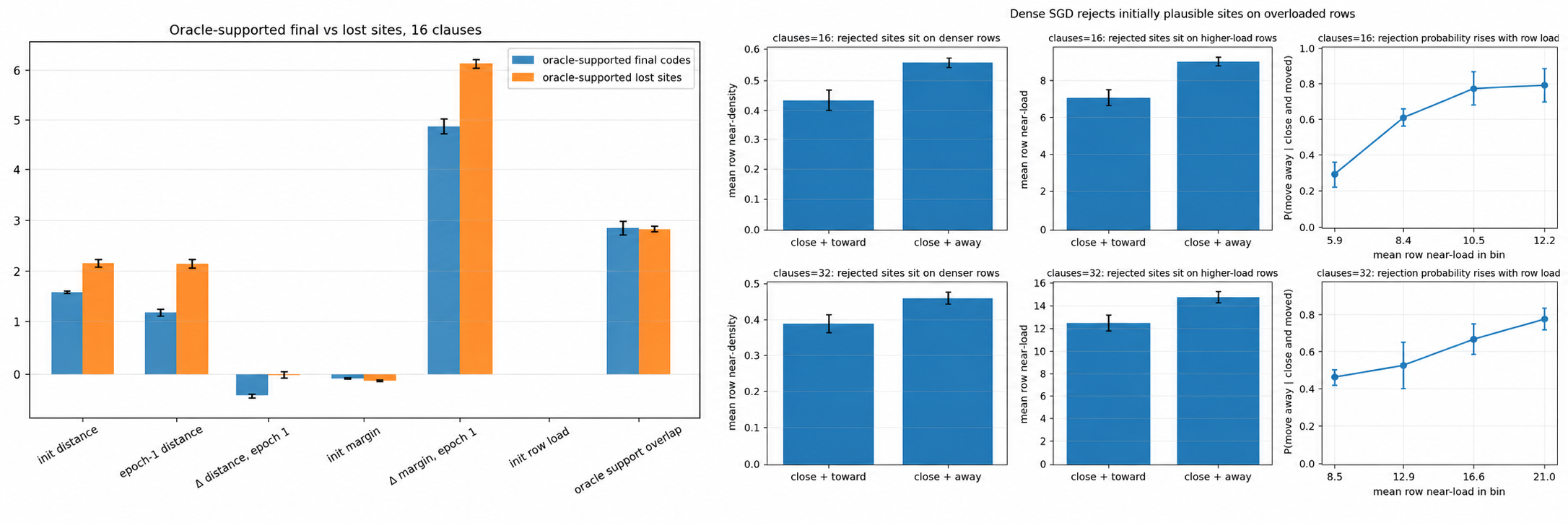}
    \caption{\textbf{Oracle-supported survival depends on both proximity and row load.} 
    }
    \label{fig:oracle-final-vs-lost}
\end{figure}

\paragraph{Row load and oracle-supported survival.}
Figure~\ref{fig:oracle-final-vs-lost} separates support membership from code survival. Within the dense-final oracle support, locations that survive as final exact codes start closer to their target family and remain closer after one epoch than supported locations that are later lost. Among initially plausible locations, rejection is concentrated on denser, higher-load rows, and the rejection probability rises with row near-load. Survival therefore depends jointly on local code geometry and row-level crowding: closer locations are more likely to become final codes, but overloaded rows reject some otherwise plausible candidates.


\section{How much structure is in a ticket?}
\label{sec:ticket-structure}

The preceding sections define a ticket as a feature-space scaffold witnessed by a sparse first-layer support.  This leaves a quantitative question open.  How much information is actually contained in that support?  Is a ticket just a random sparse expansion with the same number of parameters, or does the support encode a dense-discovered pattern of feature-space locations?

We test this by turning the usual pruning language around.  A ticket can be viewed as a \emph{sparse expansion}.  The narrow baseline is a $16\times 16$ dense $W_1$, which has $16\cdot 16=256$ active first-layer weights.  The expanded sparse model is a $32\times 16$ layer with exactly $50\%$ row-wise sparsity, so it also has $32\cdot 16\cdot 0.5=256$ active $W_1$ weights.  Both spend the same first-layer parameter budget, but the sparse expanded model has twice as many possible row carriers.

We compare three ways to spend this budget.  The first is the narrow $16\times16$ dense baseline.  The second is a random sparse expansion, in which a $32\times16$ layer receives a random $50\%$ mask at initialization and is trained with the mask fixed.  The third is a lottery sparse expansion: we train a dense $32\times16$ scout model, select a $50\%$ support using an OBS-style saliency on $W_1$, then rewind and train the sparse model with that support fixed.  We test two rewinds: to initialization and to dense epoch $10$.  We also include two references: the full $32\times16$ dense model, and a post-training OBS-compression model that keeps dense-trained weights after pruning and optionally fine-tunes them.  The latter is not a clean lottery-ticket comparison, but it is a useful upper reference for how much performance is available if the sparse model inherits the dense solution.

\subsection{Random expansion helps, but lottery expansion is more structured}
\label{subsec:random-vs-lottery-expansion}

\begin{figure}[t]
\centering
\includegraphics[width=\linewidth]{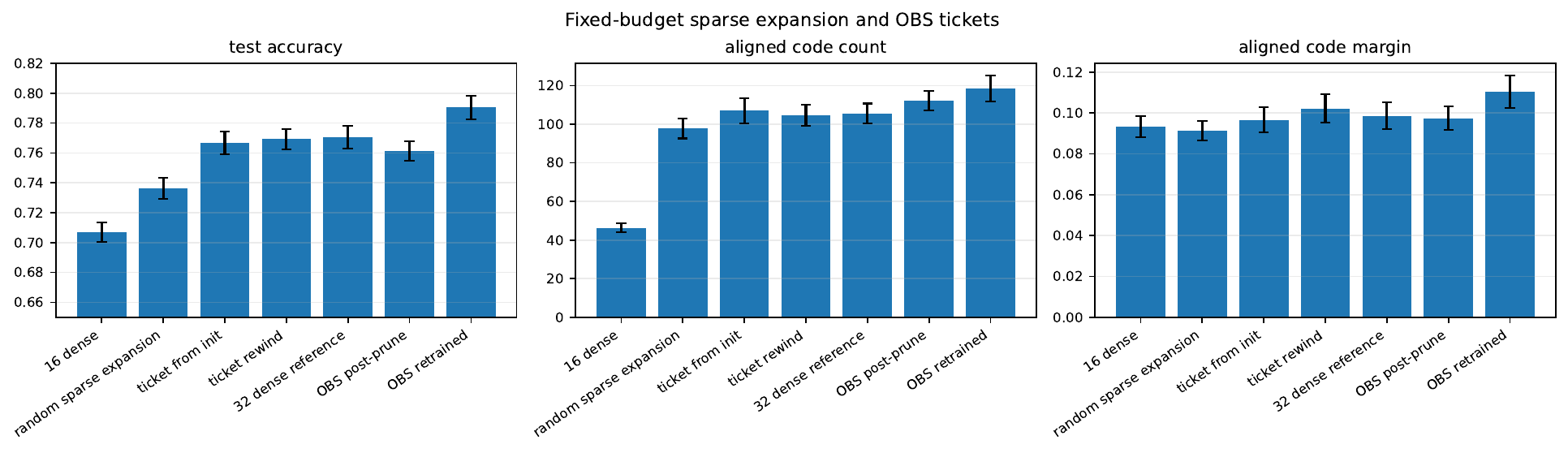}
\caption{Sparse expansion at fixed $W_1$ budget.  A random $32\times16$ $50\%$ sparse expansion improves over the $16\times16$ dense baseline, but the OBS ticket expansion is substantially better and nearly matches the full $32\times16$ dense reference.  Post-training OBS retraining is shown as a compression reference rather than as a rewound ticket.}
\label{fig:ticket-ladder}
\end{figure}

\Cref{tab:ticket-ladder} and \Cref{fig:ticket-ladder} show the basic hierarchy.  The random sparse expansion improves over the narrow dense baseline at the same $W_1$ budget: its accuracy rises from $0.707$ to $0.736$, and its aligned code count rises from $46.3$ to $97.8$.  Thus spreading the same number of first-layer parameters over more rows is useful; sparse width creates more row--clause carrier opportunities. This result was already observed in \cite{kong2025expandneuronsparameters}, and our results here simply confirm it. 

The dense-discovered OBS support is better than this random expansion.  Rewinding the OBS support to initialization gives accuracy $0.767$, and rewinding to dense epoch $10$ gives $0.769$.  These are close to the full $32\times16$ dense reference at $0.771$, even though the ticket uses only half of the dense model's active $W_1$ coordinates.  The post-training OBS model with fine-tuning reaches $0.790$, but this method inherits trained dense weights and therefore answers a different compression question.  The clean lottery comparison is the OBS ticket from init or early rewind, both of which outperform the random sparse expansion while using the same nonzero $W_1$ budget.

\begin{table}[t]
\centering
\small
\caption{Fixed-budget sparse expansion hierarchy.  All sparse $32\times16$ methods have the same active $W_1$ parameter count as the $16\times16$ dense baseline.  Values are mean $\pm$ SEM over $5$ seeds and $3$ embeddings.}
\label{tab:ticket-ladder}
\begin{tabular}{lrrrr}
\toprule
Method & $\|W_1\|_0$ & Test acc. & Aligned codes & Aligned margin \\
\midrule
16 dense & 256 & 0.707 $\pm$ 0.007 & 46.3 $\pm$ 2.3 & 0.093 $\pm$ 0.005 \\
random sparse expansion & 256 & 0.736 $\pm$ 0.007 & 97.8 $\pm$ 5.2 & 0.091 $\pm$ 0.005 \\
ticket from init & 256 & 0.767 $\pm$ 0.008 & 106.9 $\pm$ 6.6 & 0.097 $\pm$ 0.006 \\
ticket rewind & 256 & 0.769 $\pm$ 0.007 & 104.5 $\pm$ 5.5 & 0.102 $\pm$ 0.007 \\
32 dense reference & 512 & 0.771 $\pm$ 0.008 & 105.5 $\pm$ 5.2 & 0.099 $\pm$ 0.007 \\
OBS post-prune & 256 & 0.761 $\pm$ 0.006 & 112.1 $\pm$ 5.0 & 0.097 $\pm$ 0.006 \\
OBS retrained & 256 & 0.790 $\pm$ 0.008 & 118.6 $\pm$ 6.7 & 0.110 $\pm$ 0.008 \\
\bottomrule
\end{tabular}

\end{table}

This already shows that the support is not an arbitrary detail.  Random expansion gives many more codes than the narrow dense layer, but the OBS ticket gives a better functional model at the same budget.  The next question is what is structured about the OBS support.
Accuracy and aligned-code count show that the OBS support is functionally better than a random
support at the same budget. They do not yet say whether the support preserves the same
row--clause structure or merely reaches similar accuracy through a different family placement.
We therefore next compare code identities, using strict same-site overlap as a diagnostic rather than
as the definition of ticket quality.

\subsection{The ticket preserves much more of the dense code geometry}
\label{subsec:ticket-code-overlap}

We next compare the strict code identities recovered by each method.  An aligned code identity is the triple $(h,c,t)$ consisting of a row $h$, a clause $c$, and an exact row-compatible template $t$ (4P on a positive-output row or 3N1P on a negative-output row).  This is stricter than the family-level criterion used elsewhere in the paper, because it requires the same row.  We use it here as a diagnostic of how much microscopic dense geometry the mask preserves.

\Cref{tab:dense-code-recall} shows that the random sparse expansion recovers only about $20\%$ of the dense $32\times16$ aligned code identities.  The OBS ticket from initialization recovers about $53\%$, and the epoch-$10$ rewind ticket recovers about $54\%$.  The post-prune state recovers still more because it directly inherits dense-trained weights.  Thus the OBS ticket support contains more than twice as much dense code identity as the random support.

\begin{table}[t]
\centering
\small
\caption{Dense-code preservation.  Rows report strict same-site recovery of the dense $32\times16$ aligned code identities $(h,c,t)$.  The OBS tickets preserve much more of the dense scout's code geometry than a random sparse expansion.}
\label{tab:dense-code-recall}
\begin{tabular}{lrr}
\toprule
Method & Recall of dense codes & Jaccard with dense codes \\
\midrule
random sparse expansion & 0.200 $\pm$ 0.011 & 0.117 $\pm$ 0.008 \\
ticket from init & 0.528 $\pm$ 0.027 & 0.360 $\pm$ 0.023 \\
ticket rewind & 0.543 $\pm$ 0.023 & 0.377 $\pm$ 0.020 \\
OBS post-prune & 0.626 $\pm$ 0.025 & 0.442 $\pm$ 0.027 \\
OBS retrained & 0.608 $\pm$ 0.021 & 0.405 $\pm$ 0.020 \\
\bottomrule
\end{tabular}

\end{table}

\begin{figure}[t]
\centering
\includegraphics[width=\linewidth]{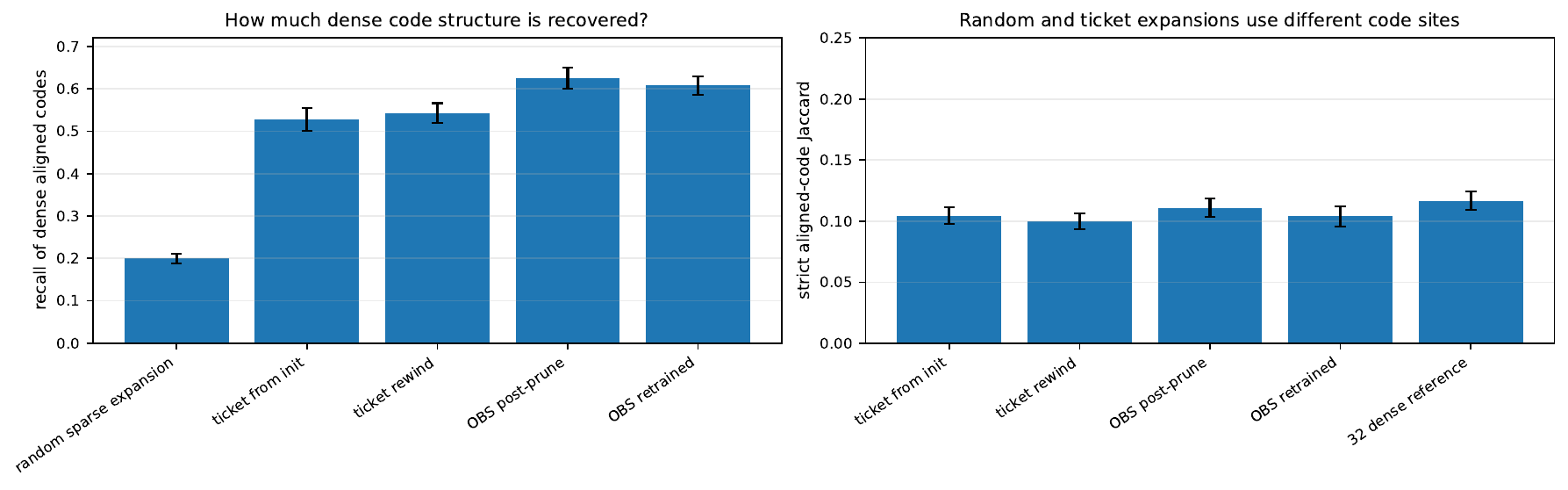}
\caption{Code-overlap diagnostics.  Left: fraction of dense $32\times16$ aligned code identities recovered by each sparse method.  Right: strict aligned-code Jaccard overlap between the random sparse expansion and other methods.  The low random-versus-ticket Jaccard shows that the OBS ticket is not merely a better-trained version of the random sparse solution; it uses a different sparse code complex.}
\label{fig:code-overlap}
\end{figure}

The random and OBS-ticket sparse models also differ from each other.  \Cref{tab:random-ticket-overlap} shows that the random sparse expansion shares only about $19$ aligned code identities with the OBS-ticket-from-init model, giving Jaccard overlap $0.105$.  The same low overlap holds for the epoch-$10$ ticket.  Therefore both models have the same width and the same number of active $W_1$ parameters, but they do not build the same code complex.  The lottery support is not just a random sparse expansion that happened to train better; it selects different row--clause sites.

\begin{table}[t]
\centering
\small
\caption{Random sparse expansion versus dense-discovered ticket supports.  The anchor is the final random $50\%$ sparse model.  Low Jaccard overlap with the OBS tickets shows that the lottery expansion is a different code complex, not simply a small perturbation of the random sparse solution.}
\label{tab:random-ticket-overlap}
\begin{tabular}{lrrr}
\toprule
Target method & Shared codes & Jaccard & Recall of random codes \\
\midrule
ticket from init & 19.2 $\pm$ 1.4 & 0.104 $\pm$ 0.007 & 0.200 $\pm$ 0.015 \\
ticket rewind & 18.1 $\pm$ 1.3 & 0.100 $\pm$ 0.006 & 0.189 $\pm$ 0.012 \\
OBS post-prune & 20.8 $\pm$ 1.5 & 0.111 $\pm$ 0.007 & 0.217 $\pm$ 0.016 \\
OBS retrained & 20.3 $\pm$ 1.8 & 0.104 $\pm$ 0.008 & 0.210 $\pm$ 0.017 \\
32 dense reference & 21.1 $\pm$ 1.7 & 0.117 $\pm$ 0.008 & 0.219 $\pm$ 0.015 \\
\bottomrule
\end{tabular}

\end{table}

\subsection{The ticket structure is already visible in the masked initial state}
\label{subsec:ticket-precursors}

The code-overlap results show that the OBS ticket ends at a different and more dense-like code complex.  We next ask whether the structure is already present before sparse training.  Applying a mask to the dense initialization changes the initial feature-space state,
\[
C_1^M(\theta_0)=(M\odot W_1^0)C_0^\top,
\]
so different supports can induce different precursor landscapes even when the surviving weights are rewound to the same dense initialization.

For every method, we measure the initial distance from each eventual final code site to its final template.  Distance $0$ means that the initial local vector is already an exact signed code; distance $1$ means a one-defect near-code.  We also measure the same quantity relative to the dense $32\times16$ final codes.  These are row-conditioned diagnostics: they ask whether the \emph{site that eventually matters} is already close to the relevant code.

\begin{figure}[t]
\centering
\includegraphics[width=\linewidth]{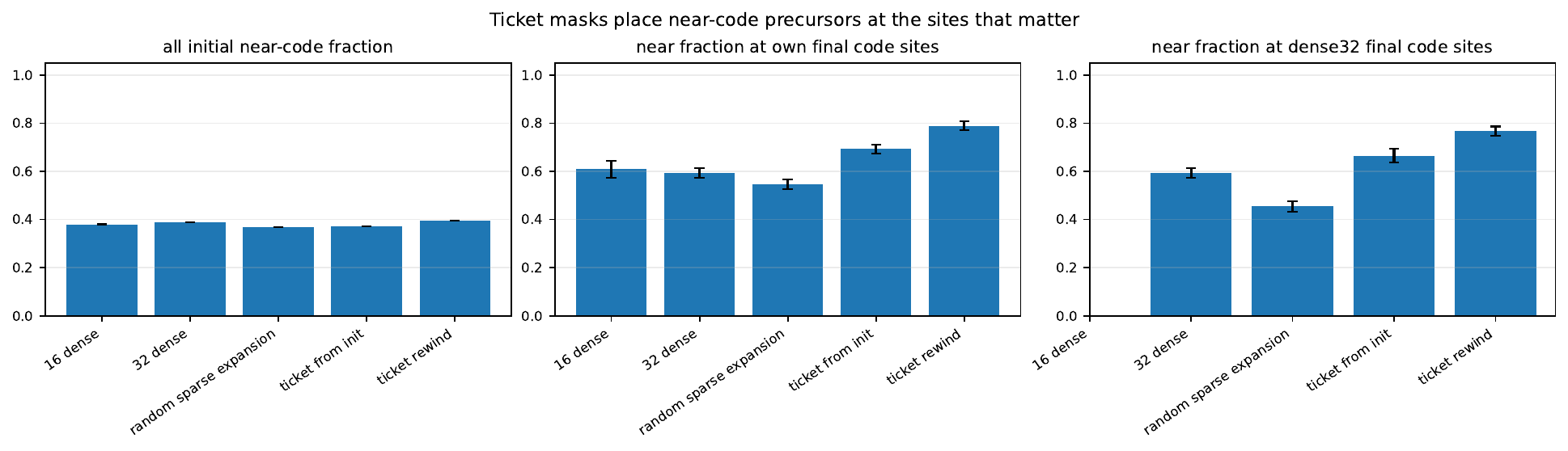}
\caption{Initial precursor structure.  Left: random and OBS masks have comparable overall density of one-defect near-codes at initialization.  Middle and right: conditioned on eventual final code sites or dense-final target sites, the OBS ticket states are much closer to the relevant templates.  This is the feature-space structure in the ticket support.}
\label{fig:precursor-summary}
\end{figure}

\Cref{tab:precursor-summary} and \Cref{fig:precursor-summary} show the key point.  Random sparse and OBS-ticket initialization have nearly the same overall density of generic near-code sites: $0.370$ versus $0.371$.  Thus the OBS support is not simply creating more near-codes everywhere.  The difference appears only after conditioning on the final code sites.  Among the sites that become the method's own final aligned codes, only $54.6\%$ of random-sparse sites are near-codes at initialization, compared with $69.3\%$ for the OBS ticket from init and $78.9\%$ for the epoch-$10$ rewind ticket.  Relative to the dense $32\times16$ final code sites, the contrast is sharper: $45.6\%$ for random sparse, $66.6\%$ for OBS-ticket-init, and $76.7\%$ for OBS-ticket-rewind.

\begin{table}[t]
\centering
\small
\caption{Initial precursor diagnostics.  ``All init near-code frac.'' scores all row--clause sites against the row-compatible family.  ``Own-final near frac.'' conditions on the sites that become each method's final codes.  ``Dense32-target near frac.'' conditions on the dense $32\times16$ final code sites.  The OBS ticket does not create more generic near-codes; it places near-codes at the sites that matter.}
\label{tab:precursor-summary}
\begin{tabular}{lrrrr}
\toprule
Method & Init near & Own near & Own dist. & Dense-target near \\
\midrule
16 dense & 0.380  & 0.609 $\pm$ 0.034 & 1.316 $\pm$ 0.065 & -- \\
32 dense & 0.389  & 0.594 $\pm$ 0.020 & 1.306 $\pm$ 0.048 & 0.594 $\pm$ 0.020 \\
random sparse & 0.370  & 0.546 $\pm$ 0.021 & 1.429 $\pm$ 0.043 & 0.456 $\pm$ 0.022 \\
ticket from init & 0.371  & 0.693 $\pm$ 0.019 & 1.127 $\pm$ 0.042 & 0.666 $\pm$ 0.030 \\
ticket rewind & 0.394  & 0.789 $\pm$ 0.018 & 0.892 $\pm$ 0.047 & 0.767 $\pm$ 0.019 \\
\bottomrule
\end{tabular}

\end{table}

\begin{figure}[t]
\centering
\includegraphics[width=\linewidth]{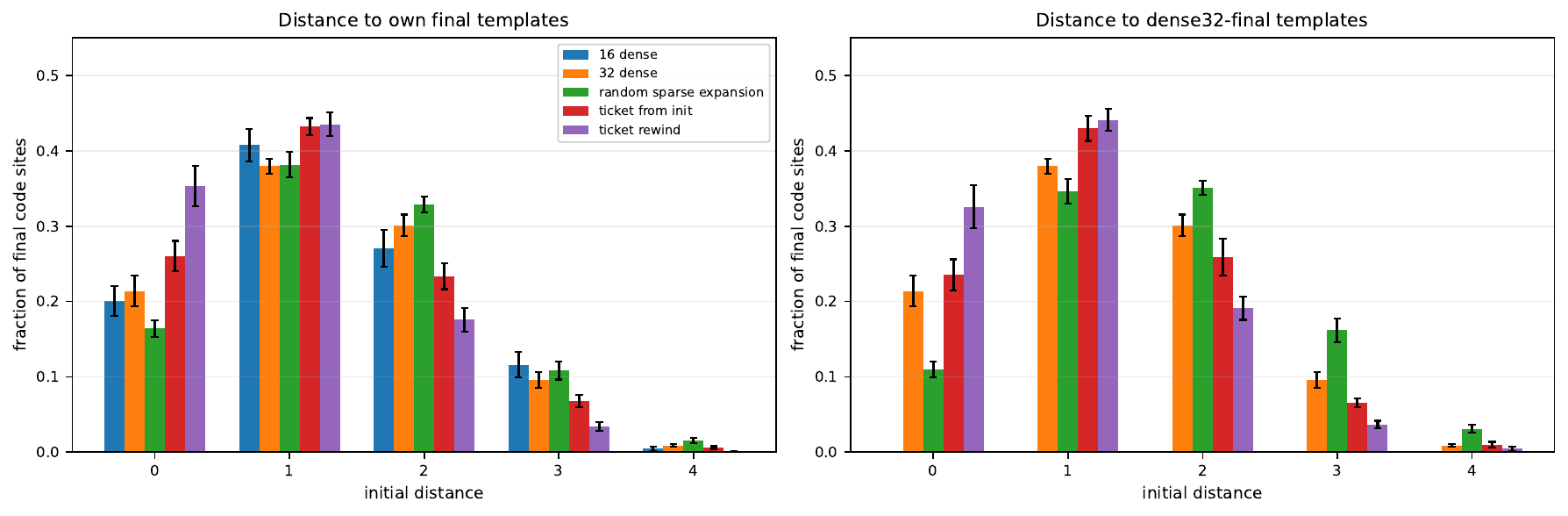}
\caption{Initial distance to final code templates.  Each bar group shows the distribution of initial distances $0,\ldots,4$ among final code sites.  The OBS tickets, especially the early-rewind ticket, place many more final code sites at distance $0$ or $1$ before sparse training begins.}
\label{fig:distance-histograms}
\end{figure}

The distance distribution in \Cref{fig:distance-histograms} gives the same result.  The OBS ticket shifts mass from distances $2$--$4$ into distances $0$--$1$.  In the epoch-$10$ rewind state, $35.4\%$ of eventual final code sites are already exact, and $78.9\%$ are within one defect.  The random sparse expansion has many eventual codes, but they start farther from their final templates.

\begin{figure}[t]
\centering
\includegraphics[width=\linewidth]{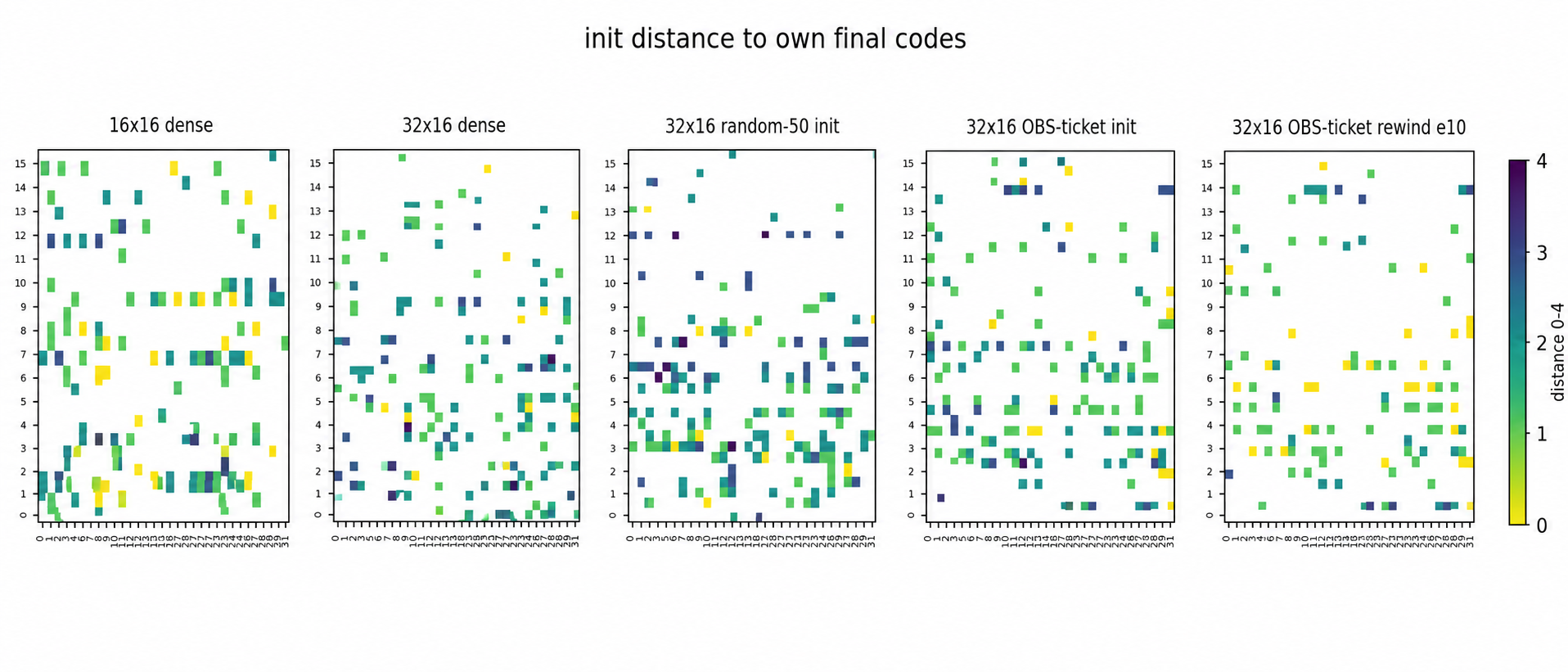}
\caption{Same-site distance from the initial local vector to the method's own final code template.  Lower values indicate better precursor placement.}
\label{fig:precursor-heatmap}
\end{figure}

\Cref{fig:precursor-heatmap} shows a representative run.  The random sparse expansion, OBS-ticket-from-init, and OBS-ticket-rewind all have the same active $W_1$ count, but their masked initial $C_1$ states differ.  The distance map in the right column marks, for each final exact code, how many sign defects the initial local vector has relative to the final template.  The OBS ticket produces visibly lower-distance precursor sites.

\subsection{What structure matters?}
\label{subsec:what-ticket-structure-matters}

These experiments separate three sources of performance.  First, sparse width itself helps: a random $32\times16$ $50\%$ expansion beats a $16\times16$ dense layer at the same $W_1$ budget.  Second, dense-discovered oracle supports help more: OBS tickets beat random sparse expansion and nearly recover the $32\times16$ dense reference.  Third, trained weights help still more: post-training OBS retraining is the best sparse method, but it is a compression baseline rather than a pure rewound ticket.

The structure in the ticket is therefore not just parameter count and not just more rows.  It has at least three measurable components.  \emph{Support geometry}: the OBS mask selects row--coordinate couplings that preserve much more of the dense scout's aligned code identities.  \emph{Targeted precursor placement}: the masked initial state induced by the OBS support is already close to the sites that become final codes, while the random mask has a similar global near-code density but worse placement.  \emph{Margin}: OBS tickets and especially the early-rewind ticket have higher aligned margins than the random sparse expansion, indicating more stable code formation.

This refines the feature-space lottery-ticket picture.  A ticket is a structured sparse expansion scaffold.  Random expansion supplies extra row carriers, but the lottery support identifies which row--coordinate directions in the expanded layer can reconstruct a compatible feature-code complex after rewind.  Pruning is the discovery procedure; the object discovered is a sparse feature-space scaffold, visible already in the masked initial state.

\section{Mechanism-inspired probes as diagnostics}
\label{sec:probe-rules}

We use mechanism-inspired mask rules as probes of the feature-space story, not as proposed general pruning algorithms. The feature-space rules first rank locations by code geometry. The static rule uses current distance and margin to the relevant row family $\mathcal T_h$; the motion rule adds finite-difference movement toward that family during dense training; the combined rule uses both. Only after ranking locations in $C_1$ do we translate them into row-wise $W_1$ masks using the contribution-aware support map described in Appendix~\ref{app:local-column}. This separation is important: finding promising code locations and realizing them with sparse $W_1$ coordinates are different problems.

\begin{figure}[th]
  \centering
  \includegraphics[width=0.97\textwidth]{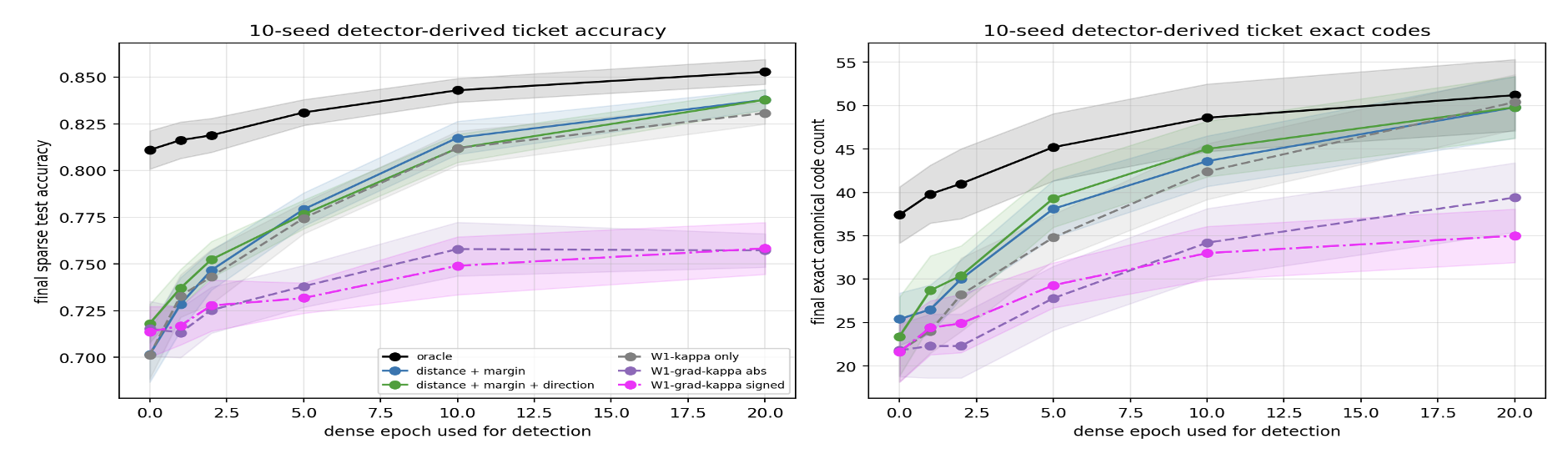}
  \caption{\textbf{Detector-derived sparse tickets under matched training conditions.}} 
  \label{fig:matched-detector-tickets-main}
\end{figure}

Figure~\ref{fig:matched-detector-tickets-main} shows the intended diagnostic contrast. Feature-space detectors are strongest where the theory predicts: early recovery of \FourP{}/\ThreeNOneP{} code structure. Later, checkpoint magnitude catches up in accuracy because dense training has already projected much of the feature-channel organization into $W_1$ magnitude order. The $W_1$-$\kappa$ and $W_1$-grad-$\kappa$ curves are weight-space controls; they score individual coordinates by coupling to clause templates, without first asking whether the corresponding location is close to a code in $C_1$. Their weaker exact-code recovery supports the claim that early ticket structure is more visible as geometry in $C_1$ than as raw weight-space saliency. Full rule definitions and broader sweeps are given in Appendix~\ref{app:probe-details}.

\begin{figure}[thb]
  \centering
  \includegraphics[width=0.98\textwidth]{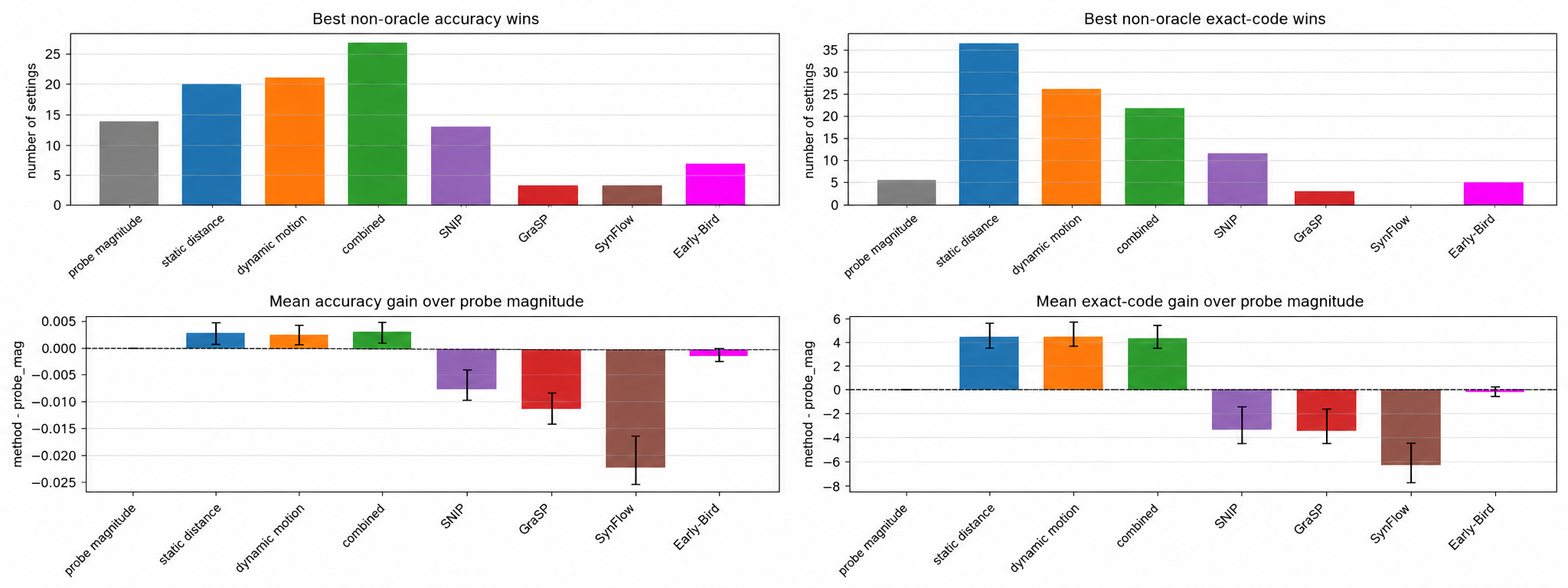}
  \caption{\textbf{Broad diagnostic sweep: feature-space masks recover the internal code structure more reliably than weight-space baselines.}}
  \label{fig:cross-setting-feature-vs-weight}
\end{figure}

\section{A broad diagnostic sweep: feature space versus weight space}
\label{sec:broad-feature-weight-sweep}

The previous section showed one matched comparison; we now use the same experiment 
as a broader diagnostic. If the ticket is a feature-space object, masks derived from 
feature-space distance and motion should recover internal code structure more reliably 
than masks chosen from raw weight-space saliency.

We sweep hidden width $H\in\{16,32\}$, clause counts $\{8,16,32\}$, $W_1$ sparsities 
$\{50\%,75\%,90\%\}$, and dense probe epochs $0,1,2,5,10,20$. At every setting each 
non-oracle rule produces a row-wise $W_1$ mask at the same sparsity budget, the sparse 
model is rewound and retrained, and we evaluate final sparse accuracy and exact canonical 
code count. Feature-space rules are static distance, dynamic motion, and a combined 
static--dynamic diagnostic; weight-space comparisons include checkpoint magnitude, SNIP, 
GraSP, SynFlow, and an Early-Bird-style magnitude rule 
\citep{lee2019snip,wang2020grasp,tanaka2020synflow,you2020earlybird}.

\begin{figure}[!htbp]
  \centering
  \includegraphics[width=0.92\textwidth]{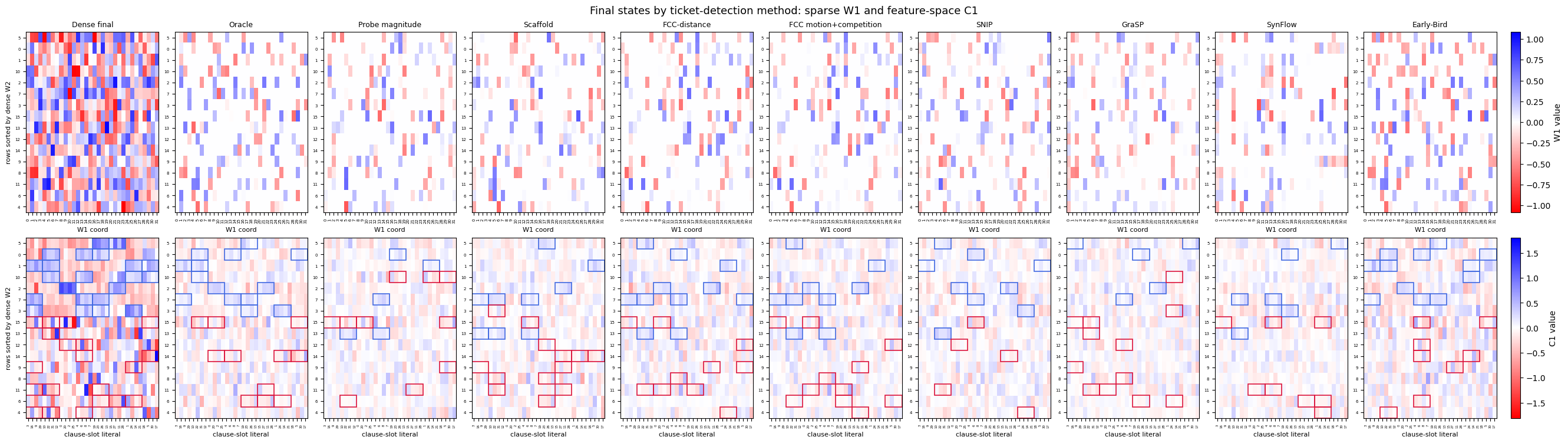}
  \caption{\textbf{Detector-derived sparse final states in \(W_1\) and \(C_1\).}
  The heat maps show final sparse representations produced by several detector rules.
  The \(W_1\) panels display the selected sparse coordinate supports after retraining,
  while the \(C_1\) panels show the clause-local feature-space computation. Exact
  \FourP{} and \ThreeNOneP{} boxes are overlaid where the final sparse model recovers
  canonical codes. The visual lesson is that detectors with similar sparsity and sometimes
  similar accuracy can produce visibly different feature-code familyes. The important
  differences are clearest in \(C_1\), not in raw \(W_1\).}
  \label{fig:app-heatmaps-detectors}
\end{figure}

Figure~\ref{fig:cross-setting-feature-vs-weight} summarizes the sweep. The three 
feature-space diagnostics win most non-oracle comparisons in both accuracy and 
exact-code recovery --- the combined rule leads on accuracy, while static distance 
and dynamic motion lead on exact-code recovery. Feature-space rules only narrowly 
beat checkpoint magnitude in accuracy, yet recover substantially more exact codes, 
a separation that matters because accuracy can stay high when a different row carries 
a family code or multiple code maps compute the same DNF\@. SNIP, GraSP, and SynFlow 
score weights or gradients rather than code locations and do not show the same 
advantage. Later checkpoint magnitude can be strong in accuracy precisely because 
dense training has already projected feature-channel organization into a $W_1$ 
magnitude ordering --- magnitude is a weight-space shadow of the structure that 
feature-space diagnostics detect earlier. One can get a feel of the differences between the various detectors by looking at the example heat maps in Figure~\ref{fig:app-heatmaps-detectors}, where each detector ends up with a different set of feature channel codes. 

The remaining failures of these probes should not be read as failures of the feature-space view alone.
They expose a second problem: translating a ranked row--clause location into a sparse set of $W_1$
coordinates. Appendix~F shows that different translation rules can preserve accuracy while changing
exact-code recovery, and Appendix~G gives the local column mechanism explaining why this happens:
a surviving coordinate contributes an entire $C_0$ column pattern to a row, so its usefulness depends
on clause-template coupling, not only on magnitude. We therefore treat the probes as diagnostics of
feature-space structure, not as final pruning algorithms.

\section{Conclusion}

We studied lottery tickets in a clause-structured setting where the relevant feature-space representation can be measured directly. In this setting, pruning is not best understood only as removing weights in $W_1$. Applying a sparse support changes the induced feature-space state $C_1 = W_1 C_0$, and sparse retraining succeeds when that new state preserves a useful scaffold of feature-channel code precursors. Dense SGD makes this scaffold legible by selectively amplifying some row--clause locations into canonical codes while rejecting others under superposition pressure. The resulting ticket is therefore not just a sparse mask, but a support-induced feature-space initialization from which sparse SGD can reconstruct a compatible family of codes.

Our experiments showed that this structure can be measured. Winning tickets preserve more useful code precursors than random sparse expansions at the same parameter budget, especially when conditioned on the sparse model's own final codes rather than only on overlap with the dense final state. The dominant task-aligned code families in the DNF setting are still \FourP{} on positive-output rows and \ThreeNOneP{} on negative-output rows, but overlapping-clause experiments also revealed secondary mixed-code structure that are not fully explained as trivial shadows of canonical codes. Mechanism-inspired probes based on feature-space distance and motion partially recover the same ticket structure and improve over raw weight-space signals for early code prediction, although the experiments also expose a remaining translation problem between selecting feature-space locations and implementing them with sparse weight-space supports.

Several directions remain open. First, the current work studies a discrete model organism in which the feature space is explicitly measurable; extending these ideas to larger networks may require approximate feature dictionaries or sparse-autoencoder-style decompositions. Second, our probe rules operate mainly at the level of canonical code families and do not yet fully capture the mixed-code structure visible in the OBS tickets. Third, the translation problem from feature-space locations to sparse weight-space coordinates remains only partially understood. Finally, while the scaling experiments suggest that the feature-space scaffold picture survives under increasing clause pressure, we do not claim that the literal 4P/3N1P alphabet or the same local combinatorial families persist unchanged in arbitrary architectures. What we expect to generalize is the broader structural phenomenon: sparse supports induce new feature-space initializations, SGD contracts selected precursor families, and lottery tickets are better understood as feature-space scaffolds than as bare masks alone.

{\small
\bibliographystyle{plainnat}
\bibliography{bibliography,refs}
}

\clearpage
\appendix
\raggedbottom
\setcounter{topnumber}{5}
\setcounter{bottomnumber}{5}
\setcounter{totalnumber}{10}
\renewcommand{\topfraction}{0.95}
\renewcommand{\bottomfraction}{0.85}
\renewcommand{\textfraction}{0.05}
\renewcommand{\floatpagefraction}{0.80}
\setlength{\textfloatsep}{8pt plus 2pt minus 2pt}
\setlength{\floatsep}{8pt plus 2pt minus 2pt}
\setlength{\intextsep}{8pt plus 2pt minus 2pt}
\setlength{\abovecaptionskip}{3pt}
\setlength{\belowcaptionskip}{0pt}

\section*{Appendix roadmap}
\label{app:roadmap}
The appendix is organized as a sequence of checks supporting the main
feature-space account. Its purpose is not to introduce a second story, but to expose the implementation choices, negative controls, and robustness checks that would otherwise interrupt the main line of the paper.
Appendix~\ref{sec:training} records implementation details.
Appendix~\ref{sec:scaling-lottery-ticket} reports scaling experiments:
accuracy and exact-code count as a function of hidden width $H$ up to
$256$, clause count, and sparsity level, supporting the claim that the
qualitative story transfers to larger settings within this model class.
Appendix~\ref{app:sparse-code-dynamics} shows that sparse retraining
contracts the locations that become sparse-final codes.
Appendix~\ref{app:family-not-site-extra} shows why family-level
recovery is a better notion than exact same-row recovery.
Appendix~\ref{app:oracle-overlap} shows that better sparse tickets
need not be literal approximations to the dense-final oracle mask,
establishing that the oracle is a conservative, basin-tied benchmark
rather than a ground truth.
Appendices~\ref{app:probe-details}--\ref{app:mask-translation} define
the detector rules and the translation from row--clause locations to
$W_1$ masks.
Appendix~\ref{app:embedding-sweeps}
tests whether the conclusions depend on the rewind basin or on the
choice of $C_0$.
Appendix~\ref{app:additional-sweeps} gives additional cross-setting
curves.

\section{Training and implementation details}
\label{sec:training}

Unless otherwise stated, experiments use binary cross-entropy loss and Adam. Dense
models are trained first. Sparse tickets are then rewound to the specified
initialization and retrained on the same data distribution while keeping the selected
\(W_1\) mask fixed. The mask is row-wise: every hidden row keeps the same fraction
of its \(W_1\) coordinates. This prevents a detector from concentrating the whole
budget in a few rows and makes the code-count comparisons interpretable across
methods.

The embedding \(C_0\) is fixed for the Hadamard and random-fixed settings and trained
jointly only in the learned-embedding condition. After dense training, the feature
space is always read through
\[
C_1 = W_1 C_0,
\]
using the paper's convention for the orientation of \(C_0\). A local row--clause
site \(C_1[h,c]\) is scored against the row-relevant family: rows with positive
output weight \(W_2[h]\) are scored against \FourP{}, while rows with negative
\(W_2[h]\) are scored against \ThreeNOneP{}. All exact-code counts use the same
threshold \(\tau\) as the main text.

Two implementation details are worth making explicit because they affect how the appendix figures should be read. First, all sparse runs keep the mask fixed throughout retraining, so any later code formation comes from changing the surviving coordinates and not from allowing pruned weights to regrow. Second, the reported code counts are feature-space diagnostics rather than additional training objectives, except in the explicitly named strengthening controls. Thus accuracy, exact-code count, and precursor distance are measured on the same trained models but answer different questions: functional success, internal code recovery, and quality of the masked initial state.

\section{Scaling of lottery-ticket benchmarks}
\label{sec:scaling-lottery-ticket}

\begin{figure}[t]
    \centering
    \includegraphics[width=\textwidth]{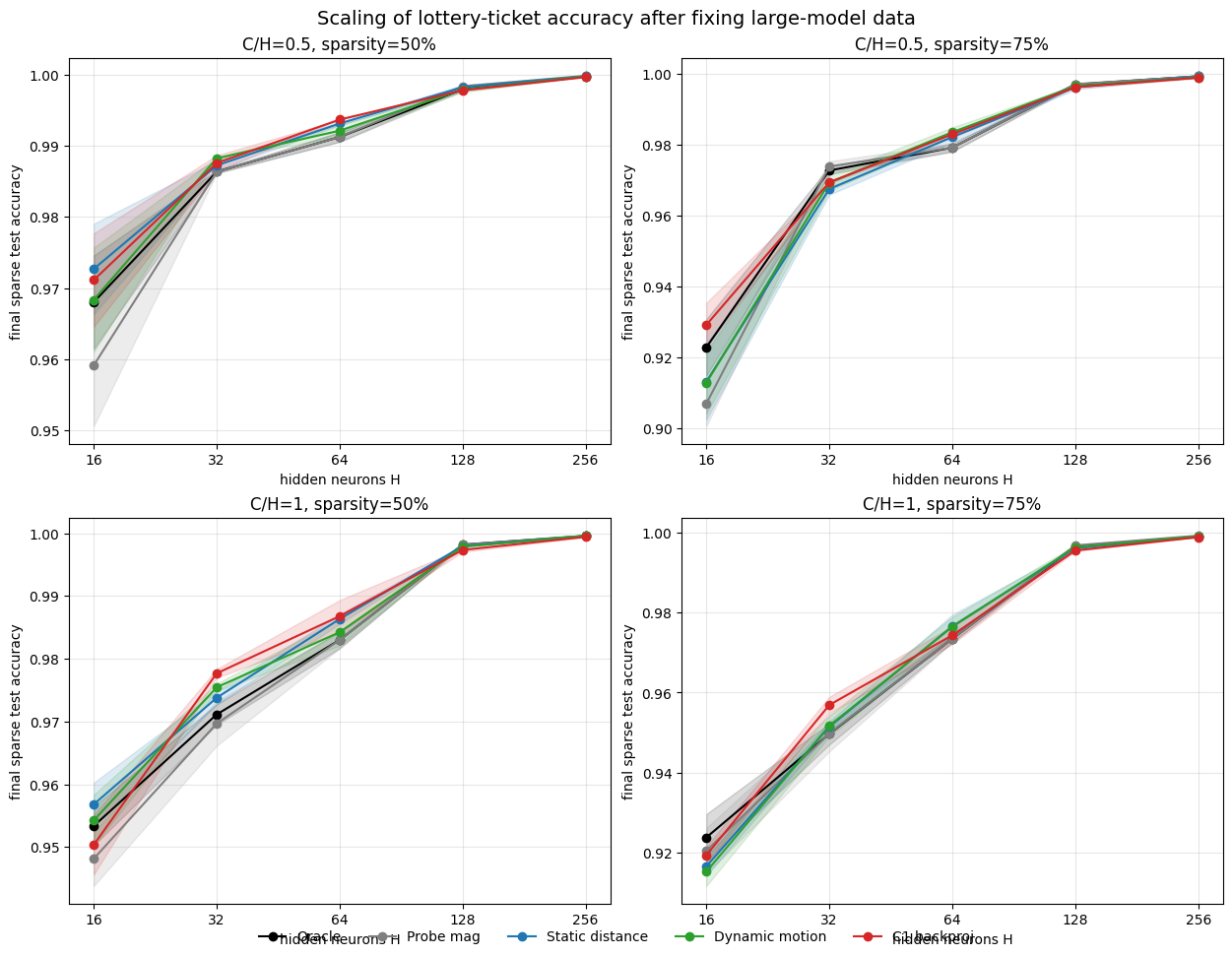}
    \caption{
    Scaling of final sparse-test accuracy after fixing the large-model data.
    Each panel varies the hidden width $H$ for a fixed class-to-hidden-width ratio $C/H$
    and sparsity level. Accuracy improves consistently with width, with the largest
    gains in the $75\%$ sparsity settings. By $H=128$--$256$, all ticket-selection
    criteria reach near-ceiling accuracy, indicating that sparse lottery-ticket
    performance is primarily width-limited rather than strongly dependent on the
    particular selection heuristic, and that the small examples used for mechanism discovery are not isolated artifacts of a single narrow setting within this model class.
    }
    \label{fig:scaling-fixed-large-model-data}
\end{figure}

\begin{figure}[t]
    \centering
    \includegraphics[width=\textwidth]{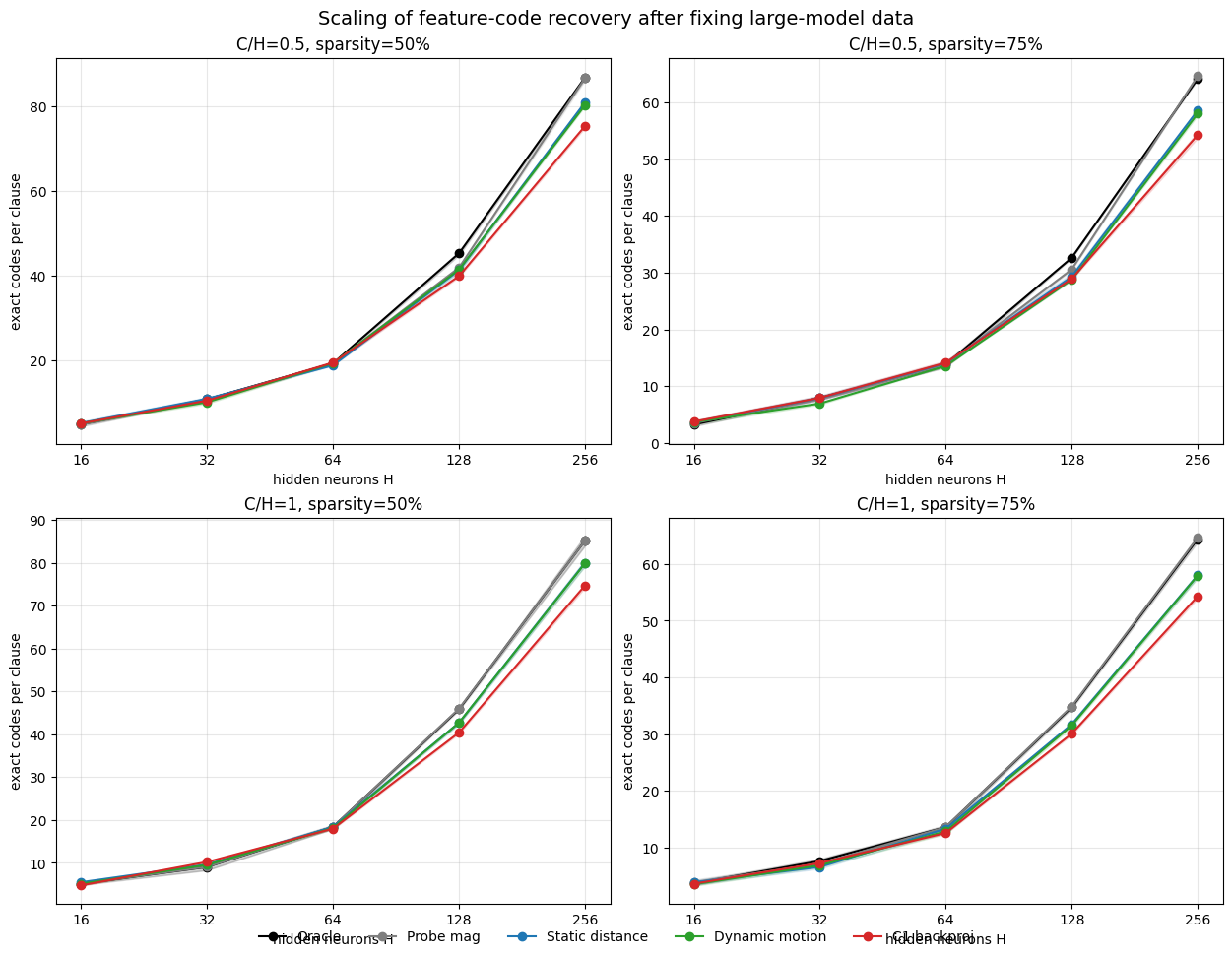}
    \caption{
    Scaling of final sparse-test code counts after fixing the large-model data.
    Each panel varies the hidden width $H$ for a fixed class-to-hidden-width ratio $C/H$
    and sparsity level. Code counts improve consistently with width. This supports the use of small models for mechanistic inspection while showing that the measured code families continue to appear in larger instances of the same toy model.
    }
    \label{fig:scaling-fixed-large-model-data-codes}
\end{figure}

We next examine how the final sparse-test accuracy scales with the hidden width $H$ after correcting the large-model data. Figure~\ref{fig:scaling-fixed-large-model-data} shows the resulting accuracies for two class-to-hidden-width ratios, $C/H \in \{0.5,1\}$, and two sparsity levels, $50\%$ and $75\%$. Across all settings, increasing $H$ improves the final sparse accuracy, with the clearest gains occurring in the more stringent $75\%$ sparsity regime. Similarly, Figure~\ref{fig:scaling-fixed-large-model-data-codes} shows that exact-code counts scale in the same direction as accuracy. 

The main scaling trend is that small networks are sensitive to both sparsity and the choice of ticket-selection criterion, whereas wider networks become substantially more robust. At $H=16$, the sparse models can incur a visible accuracy drop, especially at $75\%$ sparsity. This gap narrows quickly as $H$ increases. By $H=128$, and more clearly by $H=256$, all methods reach near-ceiling sparse accuracy across both values of $C/H$.

The comparison between selection rules also suggests that width is the dominant factor in this regime. The hidden-neuron, probe-magnitude, static-distance, dynamic-distance, and hidden-regions criteria produce broadly similar scaling curves once $H$ is moderately large. Thus, after fixing the large-model data, the qualitative conclusion is that lottery-ticket performance scales smoothly with hidden width in this toy model, and that the dependence on the specific scoring heuristic becomes small in the large-width regime. This does not mean that the detectors are equivalent: at small width and high sparsity the choice of rule still changes both accuracy and code count. Rather, the large-width setting provides enough row carriers that several reasonable rules can find a workable family scaffold.

The presented scaling experiments should be read as a stress test of the mechanism, not as a claim that the
same literal code families persist in arbitrary architectures. The robust part of the account is the
separation between weight-space support and feature-space scaffold: masks induce new initial
feature-space states, sparse retraining contracts selected precursor families, and family-level recovery
is more stable than same-row recovery. The fragile parts are the particular 4P/3N1P alphabet and the
simple site-to-weight translation rule, both of which are specific to the DNF model. Larger or less
discrete models may require learned feature dictionaries or approximate family metrics, closer in
spirit to sparse-autoencoder analyses.

\FloatBarrier

\section{Sparse retraining contracts code-distance for the sites that survive}
\label{app:sparse-code-dynamics}

The main text emphasizes dense-training trajectories. Here we show the analogous
phenomenon during sparse retraining. After the mask is selected and the surviving
weights are rewound, most locations are not exact codes. The question is whether
the locations that later become sparse-final codes are already distinguishable in
the sparse initial state and whether sparse SGD moves them toward the relevant
family. This is a self-conditioned diagnostic: locations are grouped by what happens in the sparse model itself, not by whether they match a dense-final row. It therefore tests whether the rewound sparse state already contains the precursors of its own final family locations.

\begin{figure}[!htbp]
  \centering
  \includegraphics[width=0.88\textwidth]{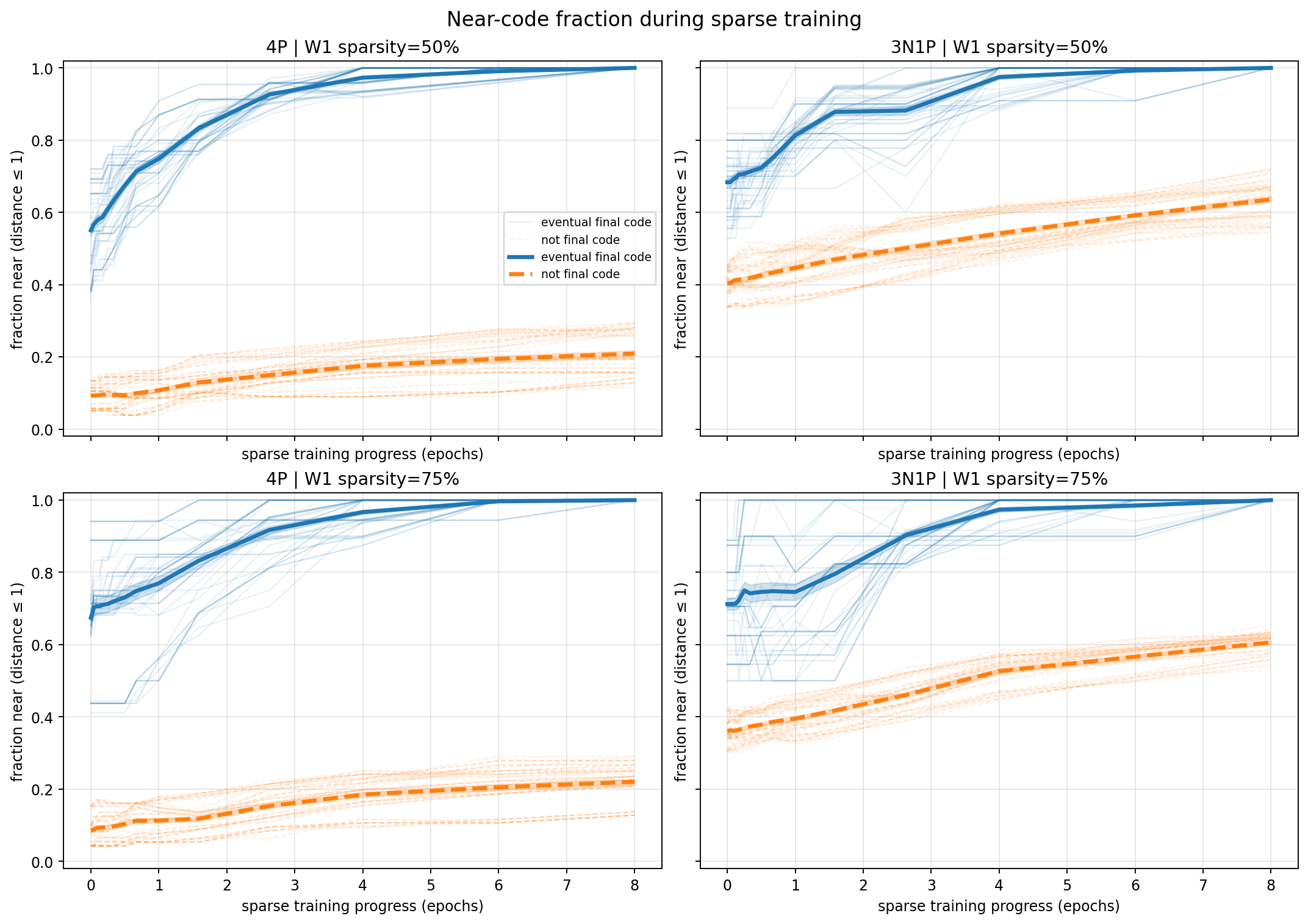}
  \caption{\textbf{Sparse retraining contracts the locations that become sparse-final codes.}
  Each panel tracks the fraction of sites that are within the near-code threshold during
  sparse retraining. The top row uses 50\% \(W_1\) sparsity and the bottom row uses
  75\% sparsity; columns separate \FourP{} and \ThreeNOneP{} families. Curves labeled
  ``eventual final code'' are the locations that become exact sparse-final codes, while
  ``not final code'' locations do not. The eventual-code locations begin closer to their
  target family and are driven toward near-code status by sparse SGD. Non-final locations
  remain much less code-like. This supports the interpretation that sparse training is
  re-expressing a feature-code scaffold rather than discovering arbitrary new structure.}
  \label{fig:app-sparse-near-code-dynamics}
\end{figure}

Figure~\ref{fig:app-sparse-near-code-dynamics} is the sparse counterpart of the dense
trajectory result. It shows that a selected support does not merely preserve final
accuracy: it creates a sparse optimization problem in which a distinguishable subset of
locations moves toward exact feature-channel codes. The separation between the blue and orange curves is important. If sparse retraining were simply discovering arbitrary new codes, the eventual-code and non-code sites would be hard to distinguish at rewind. Instead, the sites that survive start closer and contract faster, which is the operational meaning of a sparse feature-space scaffold.
\FloatBarrier

\section{Family-level recovery is stronger than same-site recovery}
\label{app:family-not-site-extra}

The main text argues that the recovered object is a feature-code family rather than a
fixed row identity. This appendix figure summarizes the same point across clause counts
and embedding families. A method can fail to recover the exact same row--clause site yet
recover the same clause/template family on another row. For a superposed representation,
that family-level recovery is the more appropriate mechanistic comparison. The row index is a carrier, not the feature identity itself. When SGD moves a clause/template pattern from one row to another, the computation can be preserved even though same-site recall records a miss.

\begin{figure}[!htbp]
  \centering
  \includegraphics[width=0.60\textwidth]{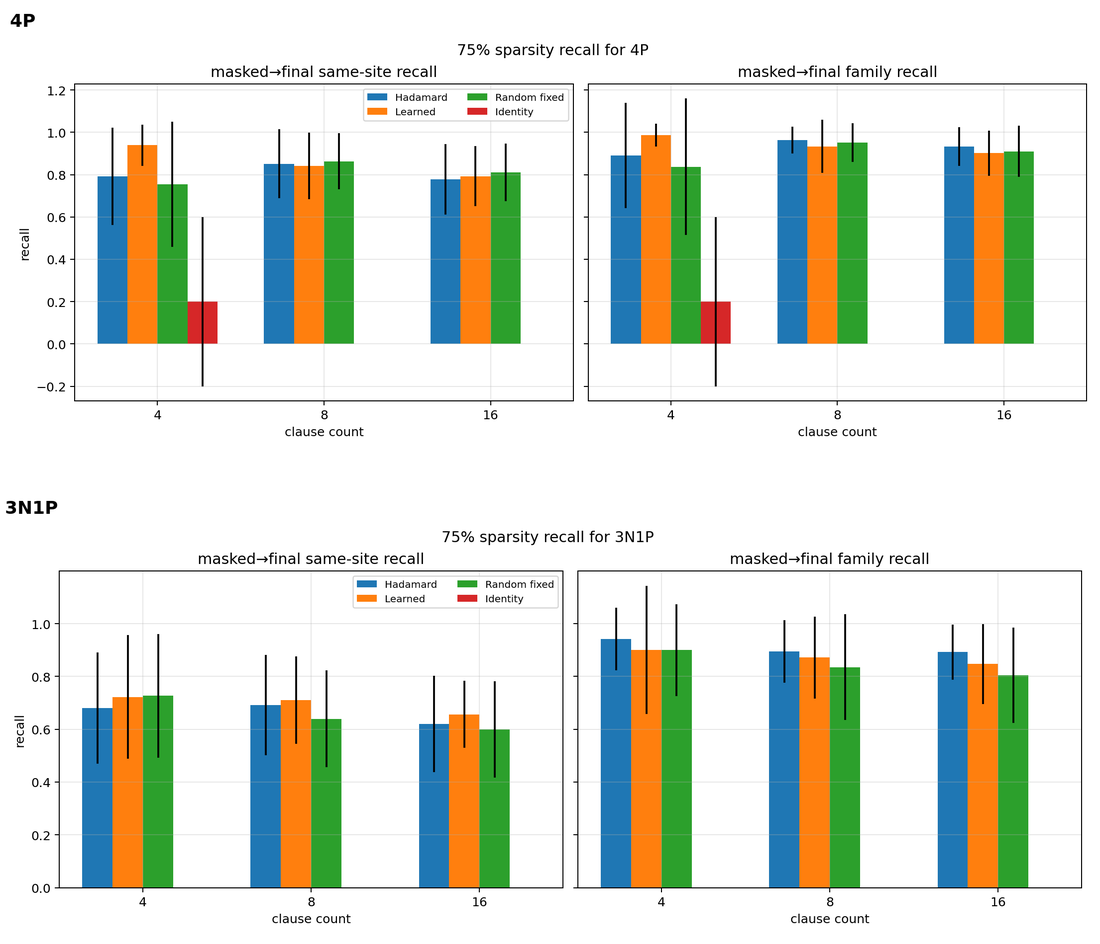}
  \caption{\textbf{Family recall is consistently higher than same-site recall.}
  The left column in each row measures exact same-site recall: whether the same
  row--clause location is preserved. The right column measures family recall:
  whether the same clause/template family is recovered somewhere in the sparse
  representation. The pattern is shown separately for \FourP{} and \ThreeNOneP{}
  codes at 75\% sparsity and across Hadamard, random-fixed, learned, and identity
  embeddings. Family recall is substantially higher than same-site recall, especially
  for larger clause counts. This is the empirical basis for treating a ticket as a
  family-level feature-code scaffold rather than a frozen set of row identities.}
  \label{fig:app-family-vs-site-recall}
\end{figure}

The implication is important for interpreting code counts. Sparse retraining can move a
clause/template family to a different row while preserving the computation. Same-site
metrics are therefore too strict for superposed codes, while family-level metrics match
the paper's claim that the ticket preserves a compatible feature-code family. This is also why the main text separates strict code-identity diagnostics from the definition of ticket quality: same-site overlap is useful for asking how much microscopic geometry was reused, but family recovery is closer to the computation realized by a superposed sparse model.
\FloatBarrier

\section{Better tickets need not imitate the oracle mask}
\label{app:oracle-overlap}

The dense-final oracle is a useful retrospective reference, but a good sparse ticket
need not match it literally in raw \(W_1\) coordinates. To test this, we compare a
feature-space rule to initialization magnitude. For a support \(M\), let
\[
J(M,M_{\mathrm{oracle}})
=
\frac{|M\cap M_{\mathrm{oracle}}|}{|M\cup M_{\mathrm{oracle}}|}
\]
be the Jaccard overlap with the dense-final oracle support. This test deliberately compares a weight-space notion of similarity to functional and feature-space outcomes. Points above the horizontal axis show that a method improved accuracy or code count; points left of the vertical axis show that it did so while matching the oracle mask less. Those points are the critical cases, because they demonstrate that literal coordinate overlap with the oracle is not the same object as the feature-space ticket.

\begin{figure}[!htbp]
  \centering
  \includegraphics[width=0.88\textwidth]{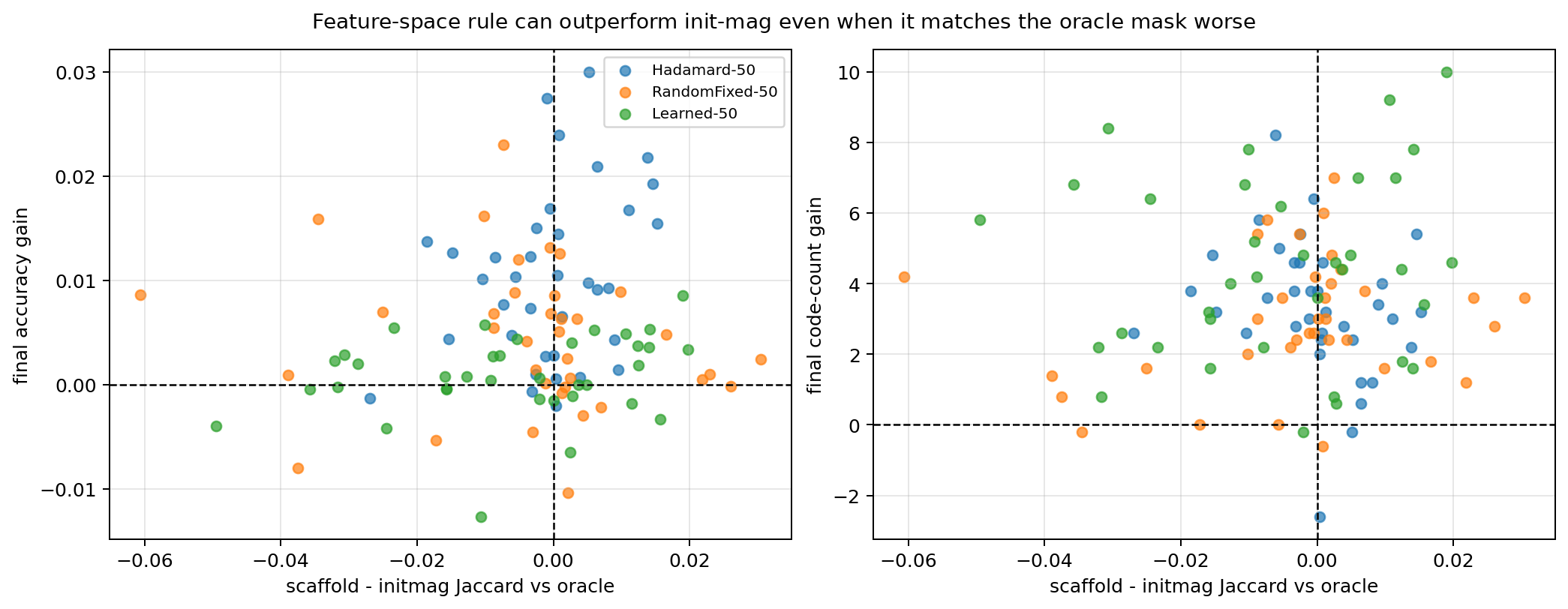}
  \caption{\textbf{Feature-space rules can improve outcomes while matching the oracle mask less.}
  Each point compares a distance-based feature-space rule to initialization magnitude in
  one embedding, clause-count, and sparsity setting. The horizontal axis is the difference
  in Jaccard overlap with the dense-final oracle support; the vertical axis is the gain in
  final sparse accuracy (left) or exact-code count (right). Many points lie above zero even
  when the feature-space rule has lower oracle-mask overlap. Thus the computation-relevant
  object is not literal coordinate overlap with the oracle. The mask is one witness to a
  feature-code scaffold, not the scaffold itself.}
  \label{fig:app-overlap-vs-outcome}
\end{figure}

Figure~\ref{fig:app-overlap-vs-outcome} explains why we do not use oracle overlap as the
primary success metric. A mask may overlap less with the oracle but still train a sparse
model with better accuracy or more exact codes. This happens because different \(W_1\)
supports can realize the same family-level computation in \(C_1\). Conversely, a support can overlap the oracle yet fail to induce the right masked initial feature-space state. The oracle is therefore best understood as a strong retrospective witness, not as a unique ground-truth mask.
\FloatBarrier

\section{Probe-rule details and baseline comparisons}
\label{app:probe-details}
\label{app:baseline-comparisons}

At dense probe epoch \(e\), let \(d_e(h,c)\) and \(m_e(h,c)\) denote the distance and
signed margin of the location \(C_1^e[h,c]\) to the relevant row family \(\mathcal T_h\).
Let \(d_0(h,c)\) and \(m_0(h,c)\) denote the corresponding values at initialization. We write
\[
\Delta d_e(h,c)=d_0(h,c)-d_e(h,c),
\qquad
\Delta m_e(h,c)=m_e(h,c)-m_0(h,c).
\]
Positive \(\Delta d_e\) means that the location moved closer to the target code family,
and positive \(\Delta m_e\) means that its signed margin improved. The directional
feature-space score used in the diagnostic rule is
\[
S_{\mathrm{dir}}(h,c)
=
z[-d_e(h,c)]
+
z[m_e(h,c)]
+
z[\Delta d_e(h,c)]
+
z[\Delta m_e(h,c)],
\]
where \(z[\cdot]\) denotes standardization over locations within the same run.

The \(W_1\)-space controls use the local column mechanism from Appendix~\ref{app:local-column}.
For a clause \(c\), template \(t\), and embedded coordinate \(j\),
\[
\kappa_{c,t,j}=\sum_{\ell\in c}t_\ell C_{0,\ell j}.
\]
The \(W_1\)-\(\kappa\) baseline scores coordinates using
\[
|W_{1,hj}\kappa_{c,t,j}|,
\]
which measures whether coordinate \(j\) has large magnitude and points row \(h\) in the
desired clause-template direction. The \(W_1\)-grad-\(\kappa\) variants additionally weight
this coupling by either gradient magnitude or signed gradient direction. These controls ask
whether the ticket is already visible directly in weight space, without first ranking
row--clause locations by code distance in \(C_1\).

The next two figures isolate the translation step. All rules start from a ranked list of sites or site-weight couplings, but they must still spend a finite row-wise budget in \(W_1\). Different translations can therefore preserve the same high-level feature-space ranking while selecting different coordinates. This is why we evaluate both accuracy and exact-code recovery: accuracy can remain stable even when the realized code family changes.

\begin{figure}[!htbp]
  \centering
  \includegraphics[width=0.90\textwidth]{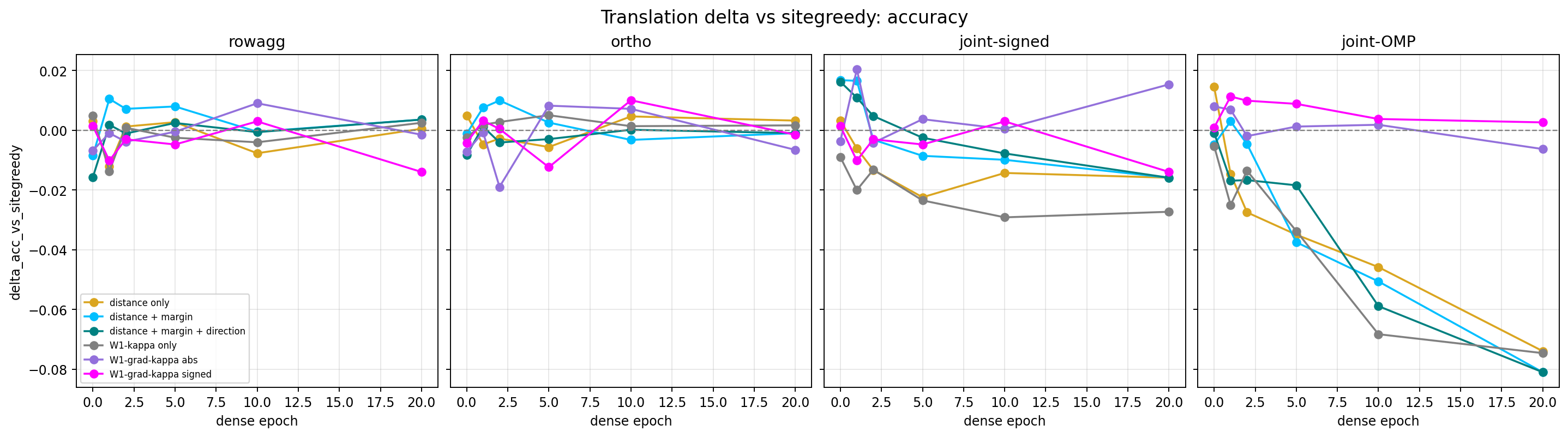}
  \caption{\textbf{Translation ablations: accuracy relative to the site-greedy translation.}
  The panels compare several ways of translating ranked row--clause sites into \(W_1\)
  masks. The vertical axis is the change in final sparse accuracy relative to a
  site-greedy conversion. Columns correspond to translation variants such as row
  aggregation, orthogonalized conversion, joint signed conversion, and joint OMP-style
  conversion. Curves correspond to the detector signals: distance only, distance plus
  margin, distance plus margin plus direction, \(W_1\)-\(\kappa\), and two gradient-weighted
  \(W_1\)-\(\kappa\) controls. The small and mixed accuracy differences show that the
  feature-space ranking and the site-to-\(W_1\) translation are separate problems: better
  site scores do not automatically imply better masks under every translation rule.}
  \label{fig:app-translation-delta-acc}
\end{figure}

\begin{figure}[!htbp]
  \centering
  \includegraphics[width=0.90\textwidth]{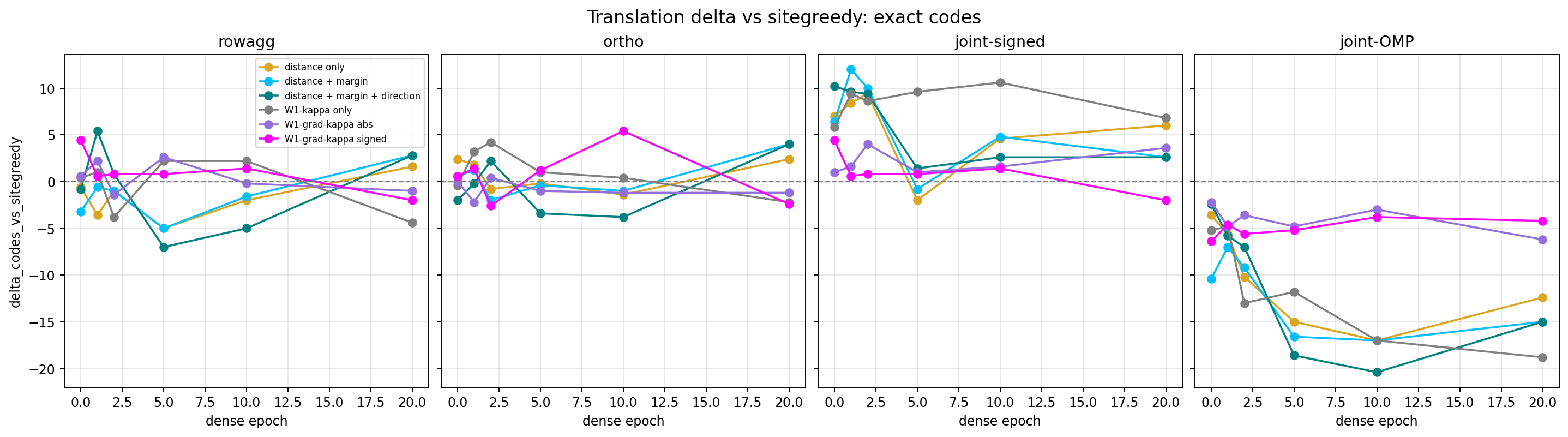}
  \caption{\textbf{Translation ablations: exact-code count relative to the site-greedy translation.}
  This figure uses the same layout as Figure~\ref{fig:app-translation-delta-acc}, but the
  outcome is the final number of exact canonical codes. The larger swings in exact-code
  count, compared with accuracy, show why code recovery is a more sensitive mechanistic
  diagnostic. Some translations preserve accuracy while losing many codes, and some
  recover more codes without large accuracy changes. This supports the main claim that
  accuracy and feature-code preservation are related but not identical objectives.}
  \label{fig:app-translation-delta-codes}
\end{figure}

Figures~\ref{fig:app-translation-delta-acc}--\ref{fig:app-heatmaps-detectors} explain why
the main paper treats detector rules as diagnostic probes rather than final algorithms. They
separate three tasks: ranking promising locations in \(C_1\), translating those locations into
a sparse \(W_1\) support, and retraining the resulting sparse model. The appendix does not claim that the site-greedy translation is optimal; it is a controlled witness that makes the feature-space ranking testable. The variation across translation rules explains why a detector can have a good site score but still lose exact codes after conversion to a mask.
\FloatBarrier

\section{The local column mechanism}
\label{app:local-column}

The clause-local matrix obeys
\begin{equation}
C_1 = W_1 C_0,
\qquad
(C_1)_{h,\ell} = \sum_j W_{1,hj} C_{0,\ell j}.
\end{equation}
A single surviving coordinate of \(W_1\) therefore contributes an entire embedding-column
pattern to a clause-local row. For a clause \(c\) and template \(t\in\{\pm 1\}^{|c|}\), define
the clause-template coupling
\begin{equation}
\kappa_{c,t,j} = \sum_{\ell\in c} t_\ell C_{0,\ell j}.
\end{equation}
Then the template score on row \(h\) can be written as
\begin{equation}
S_{h,c,t}=\sum_j W_{1,hj}\,\kappa_{c,t,j}.
\end{equation}

\begin{proposition}[Local column mechanism]
Perturbing one coordinate \(W_{1,hj}\) by \(\Delta\) changes the clause-template score by
\(\Delta\kappa_{c,t,j}\). The usefulness of a surviving coordinate is therefore controlled
not just by its magnitude but by its template coupling to the clause family.
\end{proposition}

This is the local reason raw magnitude is not enough. A coordinate matters only insofar as
it points a row in a clause-relevant direction. The same coordinate can support one clause/template family and interfere with another, because the column of \(C_0\) is reused across all clauses containing the corresponding literals. This local column view is also the bridge between the feature-space detector and the weight-space mask: after selecting a promising row--clause site, the translation rule keeps the coordinates whose columns most strongly support the desired template.

\section{From code locations to a mask}
\label{app:mask-translation}

A feature-space rule tries to approximate the oracle's ordering from the current state
rather than from the dense final model. The object it ranks is not a weight but a location.
For each row \(h\), the sign of \(W_2[h]\) determines the relevant family: rows with
positive \(W_2[h]\) are scored against \FourP{}, and rows with negative \(W_2[h]\) are
scored against \ThreeNOneP{}. For each clause \(c\) and row \(h\), we then:
\begin{enumerate}[leftmargin=1.3em]
    \item compute the best matching family template by distance and margin in the local
    vector \(C_1[h,c]\),
    \item compute a contribution-aware location support score
    \begin{equation}
    q(h,c)=\sum_{j\in\mathrm{TopK}(h,c)} |W_{1,hj}\kappa_{c,t^\star,j}|,
    \end{equation}
    where \(t^\star\) is the best local family template,
    \item rank locations lexicographically by low distance, high margin, and high \(q(h,c)\),
    \item convert the top-ranked locations into a sparse support mask by taking the
    strongest supporting weights until the row budget is filled.
\end{enumerate}
This produces a ranking of code locations and their supporting weights. Lower sparsity
means keeping more of the ranked list. The procedure is intentionally simple. It makes the detector comparison interpretable by holding the row budget fixed and using the same conversion for all feature-space rules. More sophisticated joint translations may improve performance, but they would answer a different question: how well can one optimize the mask once the feature-space site scores are known?

\section{Target-aware trajectories}
\label{app:target-aware-trajectories}

The main text classifies sites by final outcome. The target-aware diagnostic below instead
aligns sites to the final target family and asks whether the selected sites move in the
right direction during training. It provides a more direct view of the ``motion'' term used
by dynamic feature-space probes.

\begin{figure}[!htbp]
  \centering
  \includegraphics[width=0.87\textwidth]{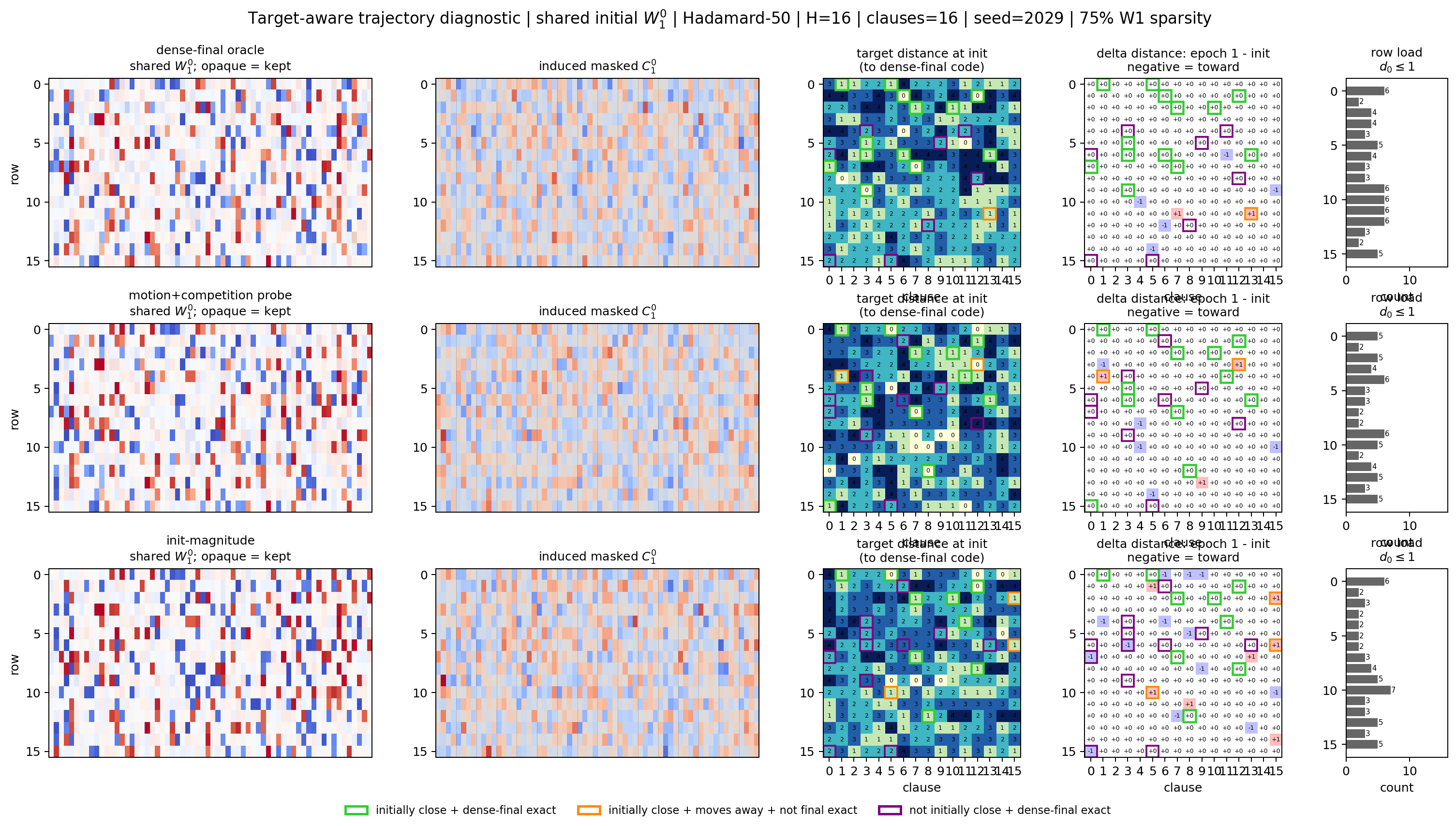}
  \caption{\textbf{Target-aware trajectories show how motion is measured.}
  Sites are grouped by their relation to the final target family and tracked over dense
  training. Curves that become final or recruited codes move toward lower distance and
  higher signed margin, while close-but-lost sites fail to maintain that improvement. This
  figure motivates the dynamic detector term \(d_0(h,c)-d_e(h,c)\): it rewards sites that
  are not merely close at one checkpoint, but are moving toward a canonical code family.}
  \label{fig:app-target-aware-trajectory}
\end{figure}

The figure clarifies why the dynamic term is not merely another way to count near-codes. A site can be close to a family at one checkpoint but then drift away or be rejected as row competition is resolved. The motion score rewards sites whose distance and margin improve in the direction of the target family, which is a stronger signal of emerging code formation than static proximity alone. Recruited codes are also visible in this diagnostic: they may not be the closest sites at initialization, but once SGD begins to allocate row carriers they move rapidly toward exactness.
\FloatBarrier

\section{Oracle support growth and embedding-family sweeps}
\label{app:embedding-sweeps}

This section checks whether the qualitative story depends on a single sparsity level, a single rewind basin, or a single embedding family. These controls are important because a dense-final oracle is tied to the trajectory that produced it. If the sparse initialization is changed, the same mask need not induce the same \(C_1\) scaffold. The first figure relaxes the oracle support from 75\% to 50\% sparsity; the second isolates the basin effect under fresh-random sparse initialization; the remaining figures repeat the dense-rewind and fresh-random comparisons across Hadamard, random-fixed, and learned \(C_0\) embeddings.

\begin{figure}[!htbp]
\centering
\includegraphics[width=0.88\linewidth]{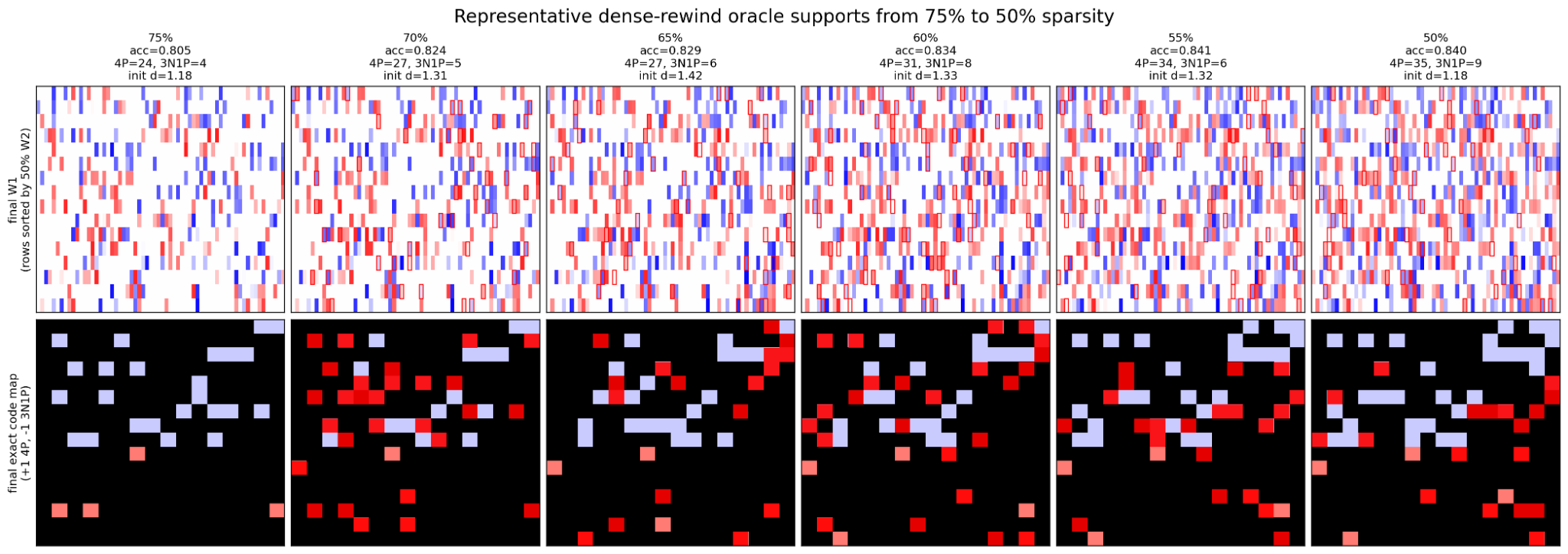}
\caption{\textbf{Oracle support growth from 75\% to 50\% sparsity.}
Looser supports add code locations rather than arbitrary isolated weights. The sparse
oracle's feature-space support grows in a structured way: additional retained coordinates
bring in more clause-local code sites. This confirms that the main 75\% examples are not
isolated artifacts of a particular sparsity budget.}
\label{fig:app-oracle-sparsity}
\end{figure}

\begin{figure}[!htbp]
\centering
\includegraphics[width=0.88\linewidth]{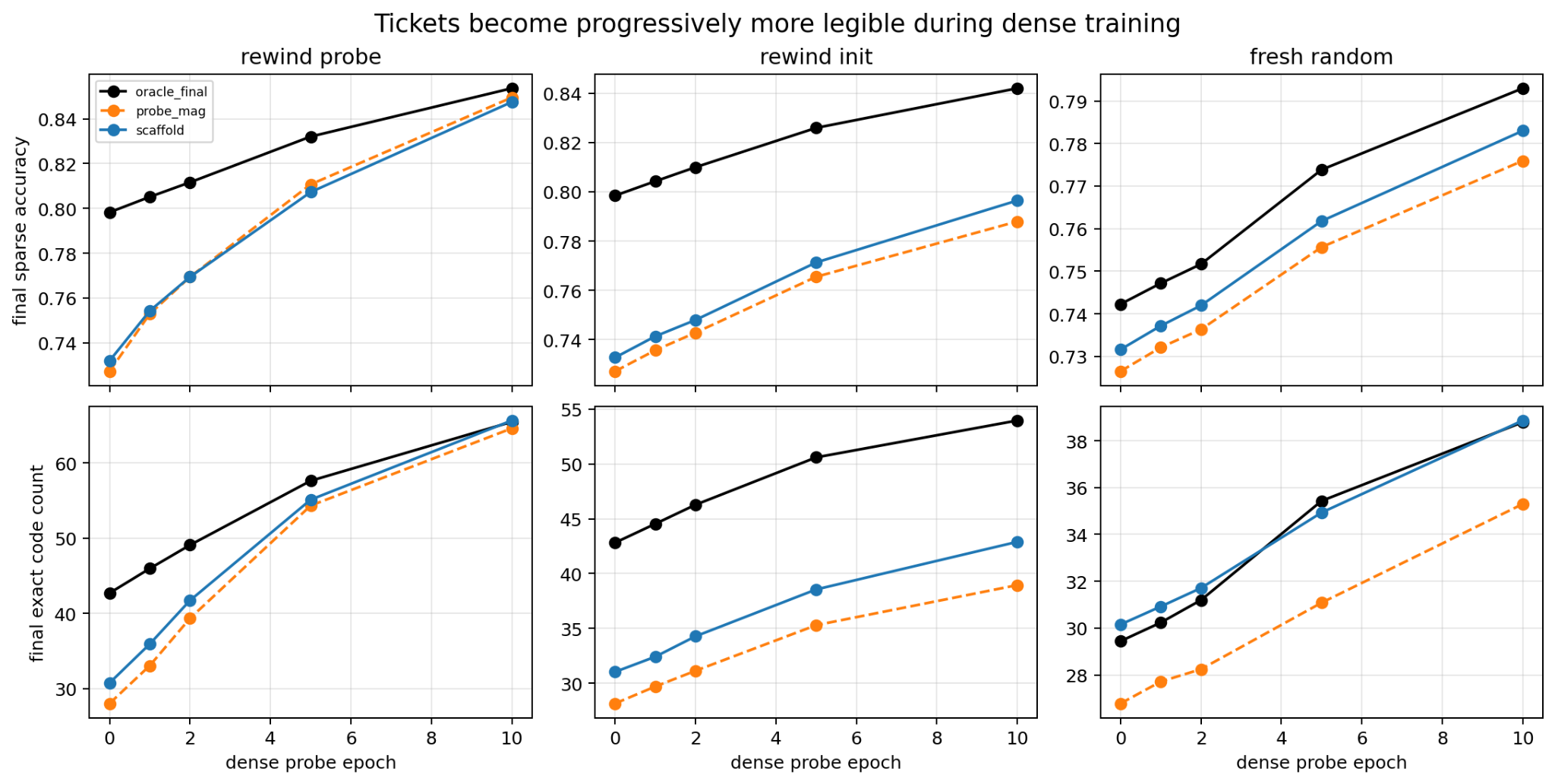}
\caption{\textbf{Fresh-random initialization shows why the dense-final oracle is basin-tied.}
Columns correspond to sparse initialization modes: rewind to the dense probe checkpoint,
rewind to the original dense initialization, and fresh-random initialization. The top row
reports final sparse accuracy; the bottom row reports exact canonical code count. Under
matched rewind settings, the dense-final oracle remains the strongest retrospective support.
Under fresh-random initialization, feature-space rules become more competitive because they
rank the current sparse-initial geometry rather than copying the final support of another
trajectory. The result supports the view that a ticket is not a mask alone, but a support
paired with the feature-space state induced by applying that support at rewind.}
\label{fig:app-fresh-random-state-adaptive}
\end{figure}

The fresh-random comparison is a negative control on the idea that a mask has an intrinsic value independent of the state in which it is used. The oracle mask is excellent when it is paired with its own rewind basin, but it loses part of that advantage when the sparse initial coefficients are redrawn. State-adaptive feature-space rules can then recover more of the current code geometry because they look at the present \(C_1\) state rather than at another trajectory's dense-final weights.

\begin{figure}[!htbp]
\centering
\begin{subfigure}{0.43\linewidth}
\includegraphics[width=\linewidth]{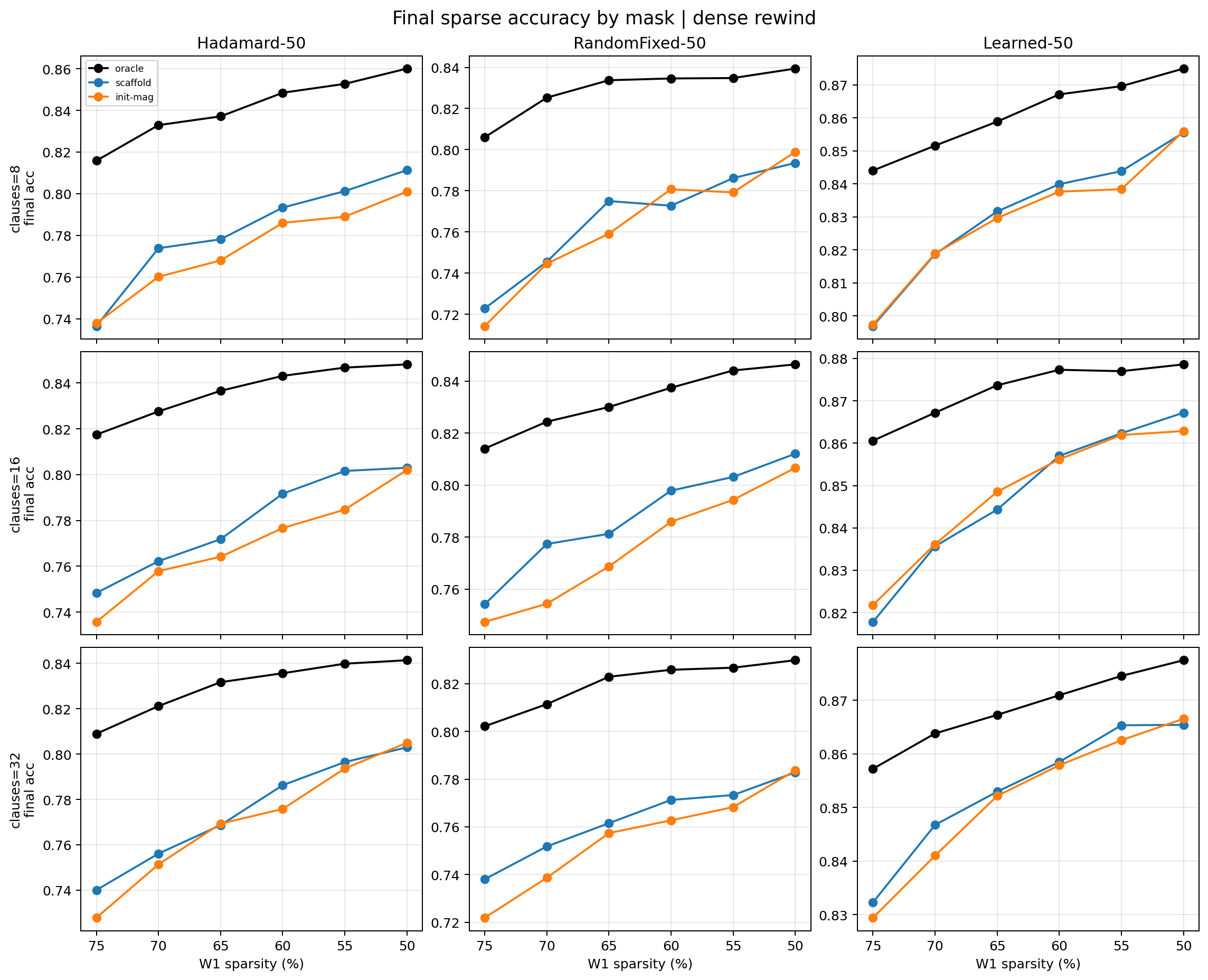}
\caption{Accuracy under dense rewind.}
\end{subfigure}
\begin{subfigure}{0.43\linewidth}
\includegraphics[width=\linewidth]{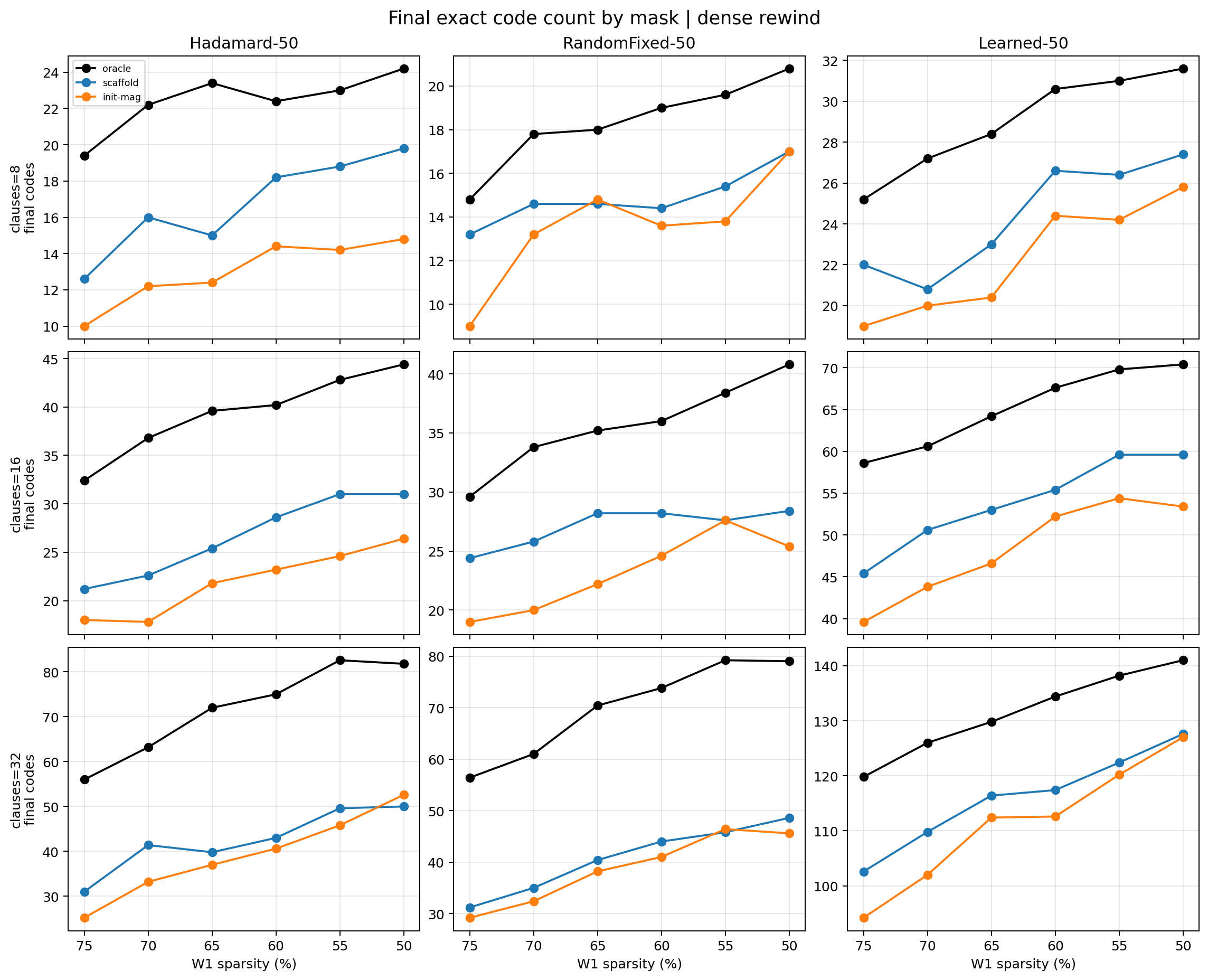}
\caption{Exact codes under dense rewind.}
\end{subfigure}
\caption{\textbf{Embedding-family sweep under dense rewind.}
Hadamard-50, RandomFixed-50, and Learned-50 embeddings show the same qualitative ordering:
feature-space detectors remain competitive across embeddings, and code recovery is a more
sensitive diagnostic than accuracy. The result argues that the feature-space ticket story is
not specific to a Hadamard \(C_0\).}
\label{fig:app-embedding-dense-rewind}
\end{figure}

Figure~\ref{fig:app-embedding-dense-rewind} shows that the dense-rewind setting is not an artifact of a single embedding. The learned, random-fixed, and Hadamard embeddings differ in how the column patterns of \(C_0\) distribute literal information, but the same qualitative comparison persists: rules that read feature-space distance and motion remain competitive, and exact-code recovery is more sensitive than accuracy.

\begin{figure}[!htbp]
\centering
\begin{subfigure}{0.70\linewidth}
\centering
\includegraphics[width=\linewidth]{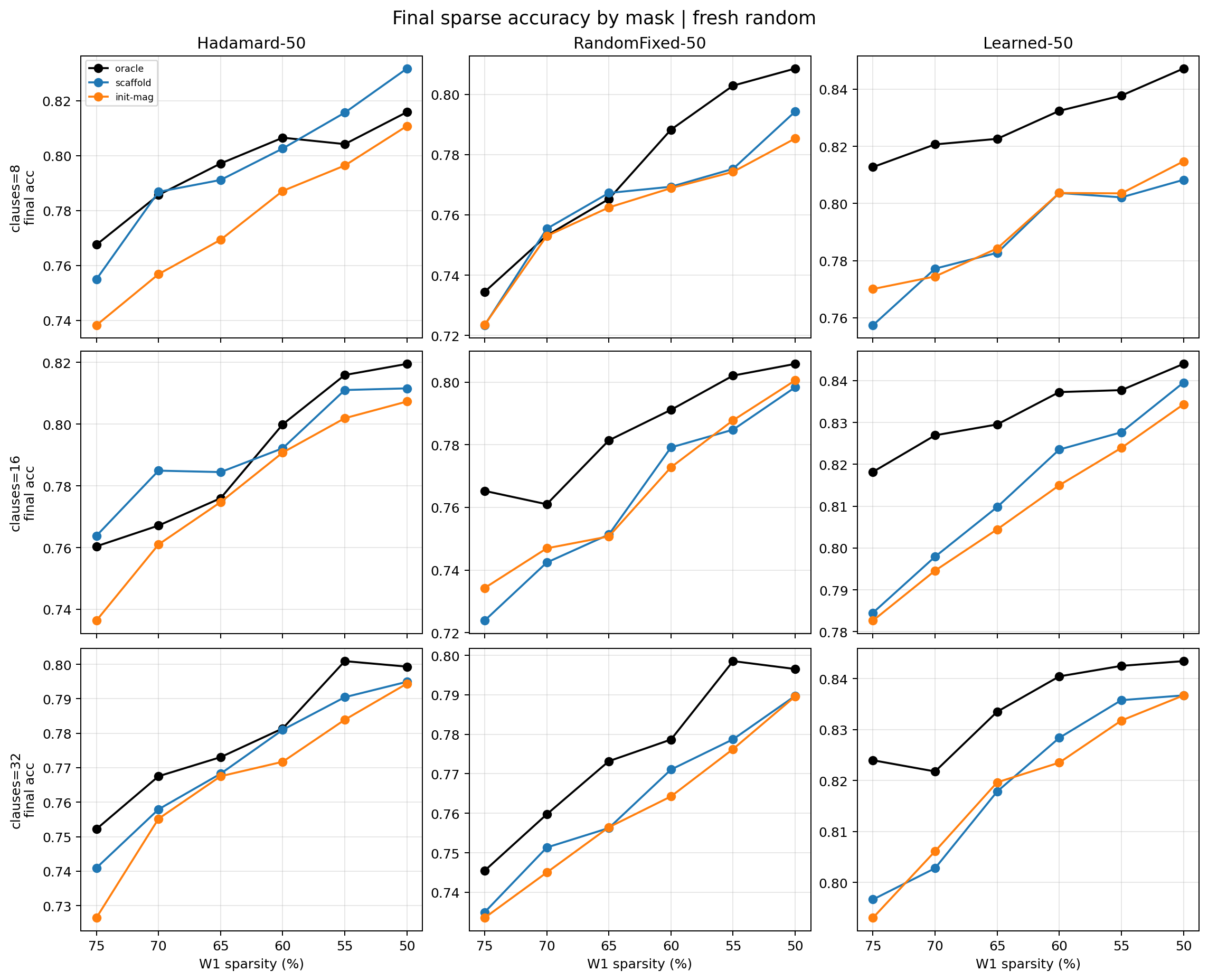}
\caption{Accuracy under fresh-random sparse initialization.}
\end{subfigure}
\begin{subfigure}{0.92\linewidth}
\centering
\includegraphics[width=\linewidth]{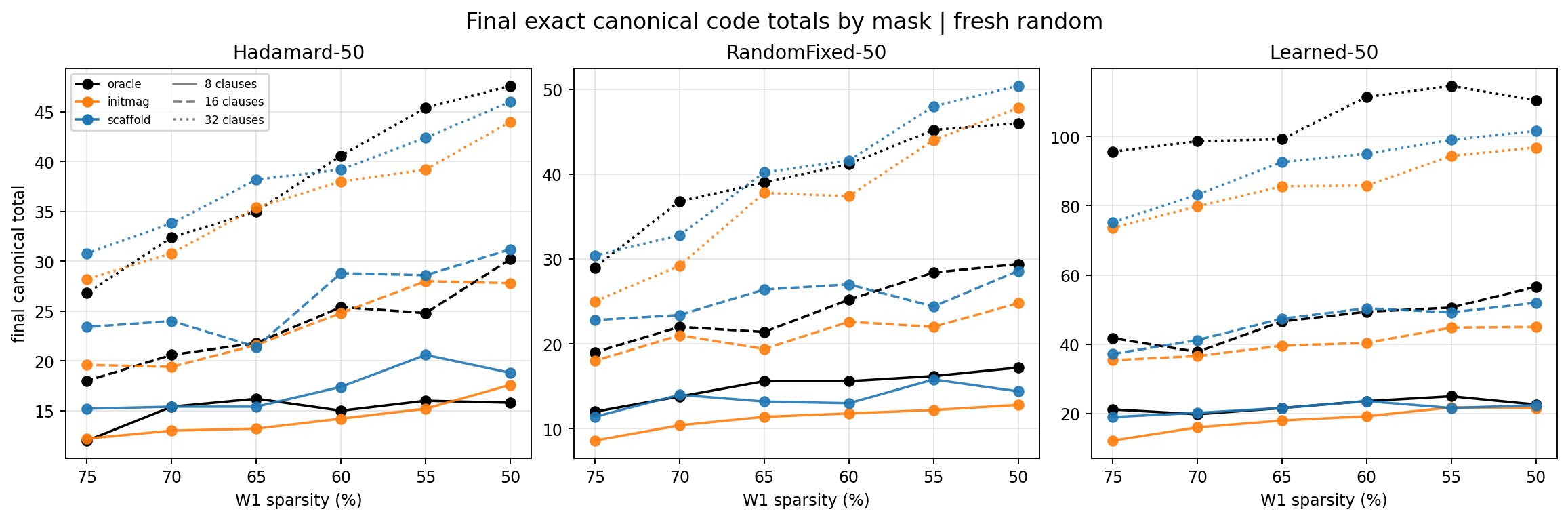}
\caption{Exact codes under fresh-random sparse initialization.}
\end{subfigure}
\caption{\textbf{Embedding-family sweep under fresh-random sparse initialization.}
When the sparse initialization no longer matches the dense-final oracle basin, the
state-adaptive feature-space rules become more competitive across embeddings. The exact-code
panels show the clearest version of this effect: rules that read the current feature-space
geometry can recover code structure even when the dense-final oracle loses part of its basin
advantage.}
\label{fig:app-embedding-fresh-random}
\end{figure}

Figure~\ref{fig:app-embedding-fresh-random} repeats the same stress test across embeddings. The key pattern is the same as in Figure~\ref{fig:app-fresh-random-state-adaptive}: once the sparse model is no longer initialized in the oracle's own basin, state-adaptive feature-space rules become more competitive. This reinforces the distinction between a mask and a feature-space ticket witness. The mask is useful because of the scaffold it induces in a particular state, not because its coordinate set is universally privileged.
\FloatBarrier

\section{Additional cross-setting curves}
\label{app:additional-sweeps}

The main text includes the compact aggregate from the cross-setting sweep. The figure below
shows representative time courses for \(H=16\), clause counts 8 and 16, and several sparsity
levels. These curves provide the underlying view behind the aggregate win counts. They also illustrate why we report both accuracy and exact-code count. Some rules become competitive in accuracy as the dense probe epoch increases, while exact-code recovery can still separate feature-space and weight-space detectors. The gap is informative: accuracy measures whether the DNF is solved, whereas exact-code count measures whether the detector recovered the intended internal family scaffold.

\begin{figure}[!htbp]
\centering
\includegraphics[width=0.86\linewidth]{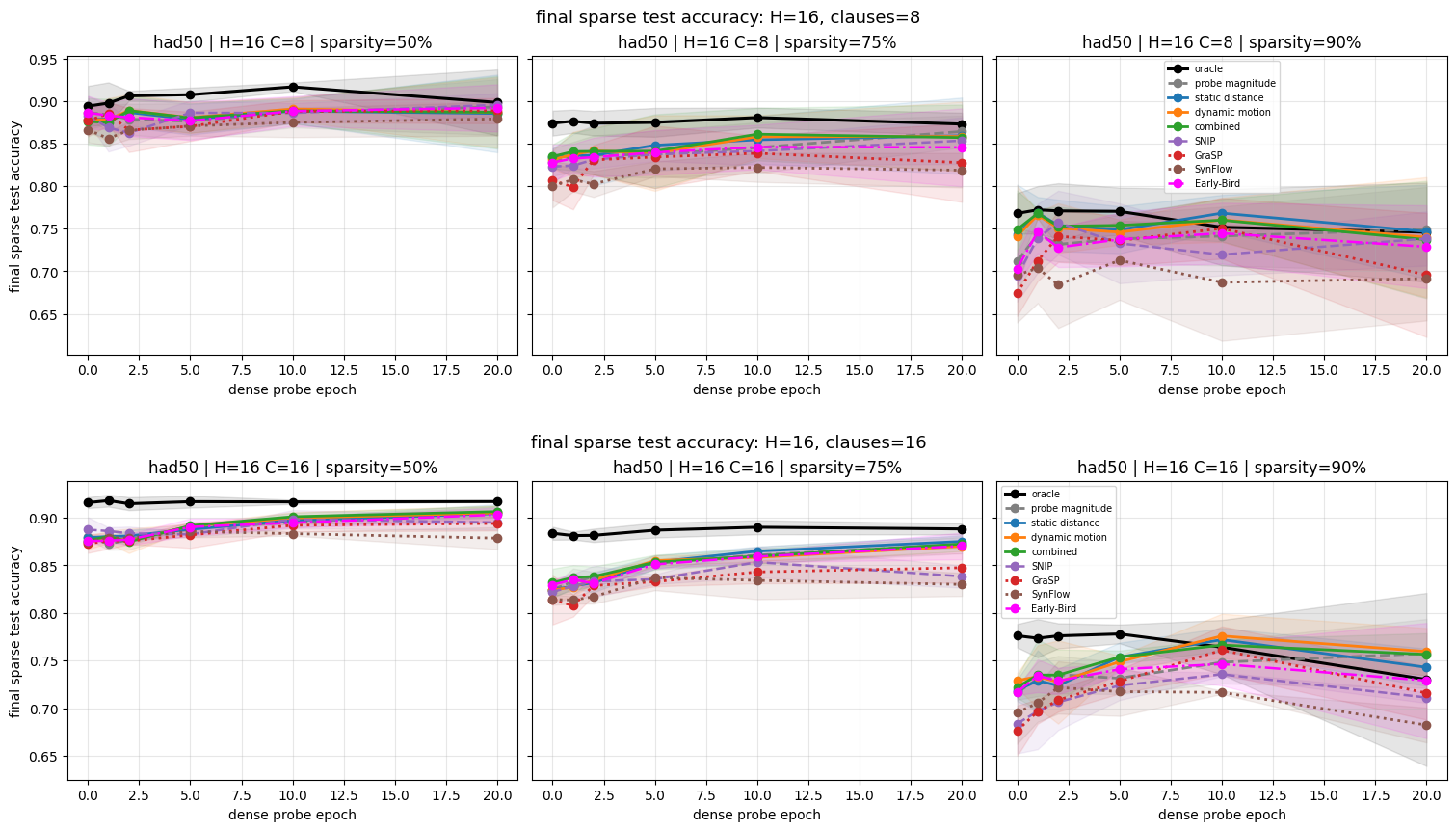}
\caption{\textbf{Representative cross-setting accuracy curves.}
The panels show final sparse accuracy as a function of the dense epoch used for detection,
for \(H=16\), clause counts 8 and 16, and multiple sparsities. Feature-space diagnostics are
competitive early, while checkpoint magnitude becomes a strong accuracy proxy later in
training. This supports the interpretation that dense training gradually projects the
feature-code scaffold into the \(W_1\) magnitude ordering.}
\label{fig:app-cross-setting-curves}
\end{figure}

Together, the appendix figures support the same division of labor used in the main text. Feature-space distance and motion are early indicators of the family scaffold; weight magnitude becomes informative later, after dense training has already organized the relevant computation; and mask translation remains a separate source of variation. The appendix therefore strengthens rather than changes the main claim: lottery tickets in this model are best understood as support-induced feature-space family scaffolds witnessed by masks.

\end{document}